\documentclass[11pt]{article}

\usepackage[preprint]{acl}
\usepackage{times}
\usepackage{latexsym}
\usepackage{booktabs}
\usepackage{multirow}
\usepackage{colortbl}
\usepackage{makecell}
\usepackage{arydshln}
\usepackage{amsmath}
\usepackage{amssymb}
\usepackage{amsfonts}
\usepackage{tikz} 
\definecolor{best1}{HTML}{D7EAFB}
\definecolor{best2}{HTML}{E6F4EA}
\usepackage{hyperref}

\newcommand{\modf}[1]{%
  \tikz[baseline=-0.6ex]\node[
    circle, 
    fill=gray!40, 
    inner sep=0pt,
    minimum size=4mm,
    font=\sffamily\scriptsize\bfseries
  ] {#1};%
}

\newcommand{\mode}[1]{%
  \tikz[baseline=-0.6ex]\node[
    circle, 
    draw, 
    dash pattern=on 1pt off 1pt, 
    line width=0.4pt, 
    text=gray, 
    inner sep=0pt, 
    minimum size=4mm, 
    font=\sffamily\scriptsize
  ] {#1};%
}

\newcommand{\mc}[3]{%
  \ifnum#1=1 \modf{A}\else\mode{A}\fi\hspace{1pt}%
  \ifnum#2=1 \modf{T}\else\mode{T}\fi\hspace{1pt}%
  \ifnum#3=1 \modf{V}\else\mode{V}\fi
}

\usepackage[T1]{fontenc}
\usepackage[utf8]{inputenc}
\usepackage{microtype}
\usepackage{inconsolata}
\usepackage{graphicx}

\title{Modality Disentangled Learning for Incomplete Multimodal Emotion Recognition: A Primitive Memory Distillation Perspective}

\author{
  \textbf{Jiaqi Zhang\textsuperscript{1}},
  \textbf{Zheng Pang\textsuperscript{1}},
  \textbf{Mengting Li\textsuperscript{1}},
  \textbf{Yiqi Wang\textsuperscript{2}},
  \textbf{Guangyuan Dong\textsuperscript{3}\thanks{Equal Contribution with the First Author.}},
  \textbf{Chao Xue\textsuperscript{4}},
  \\
  \textbf{Yusen Wu\textsuperscript{5}},
  \textbf{Zihao Li\textsuperscript{6}},
  \textbf{Huy Phan\textsuperscript{7}},
  \textbf{Sicheng Zhao\textsuperscript{8}},
  \textbf{ Bj\"orn~W.~Schuller\textsuperscript{9,10}},
  \textbf{Jiachen Luo\textsuperscript{9,11}\thanks{Corresponding Author. {jiachen.luo@qmul.ac.uk}}}
\\
\\
  \textsuperscript{1}Jiangsu University
  \textsuperscript{2}Griffith University
  \textsuperscript{3}National University of Singapore\\
  \textsuperscript{4}University of New South Wales
  \textsuperscript{5}Fujian University of Technology \\
  \textsuperscript{6}Xi'an Jiaotong-Liverpool University
  \textsuperscript{7}German Research Center for Artificial Intelligence \\
  \textsuperscript{8}Tsinghua University
  \textsuperscript{9}Technical University of Munich\\
  \textsuperscript{10}Imperial College London
  \textsuperscript{11}Queen Mary University of London
}

\begin{document}
%全局术语一致性（做一次全文替换）！！！
% 保留	删除/替换掉
% full-modality	complete-modality, complete modality
% missing-modality (形容词)	modality-missing
% available modalities	visible modalities, observed modalities
% disentangle	decouple（除 SSSD 缩写本身外）
% semantic primitives	semantic prototypes
% student model / teacher model	student / teacher（单独出现时补上 model）

% 保留	替换掉
% primitive	prototype（出现在 hyperparameter sensitivity、ablation、SOTA 讨论等 ≥6 处）
% available modalities	visible modalities / observed modalities
% full-modality	complete-modality
% missing-modality（形容词）	modality-missing
% Retrieval-Augmented	Retrieval Augmented
% L_{\mathrm{Align}}	L_{Align}（Table 3 与正文用了无 \mathrm 版本，与公式不一致）
% \tau_{\mathrm{KD}}	\tau_{KD}（超参表）
% \operatorname{Concat}	\operatorname{Concat} 混用

\maketitle
\begin{abstract}
Multimodal Emotion Recognition (MER) systems often suffer from missing modalities in real-world scenarios. Existing methods usually generate, align, or distill missing modalities as a whole, overlooking the heterogeneous nature of the information carried by each modality. Such holistic treatment mixes inferable shared semantics with uncertain modality-specific details, yielding unstable representations and degrading robustness. To address this issue, we propose the Primitive Memory Distillation (PriMD) framework. Unlike existing methods, PriMD takes an intra-modal perspective and focuses on how different types of information within a modality differ in recoverability. PriMD first disentangles cross-modal shared semantics from modality-specific representations, and then discretizes the latter into learnable semantic primitives to construct modality-specific memory banks. When modalities are missing, PriMD is a teacher–student framework that the student model uses the shared semantics of available modalities as queries to dynamically retrieve primitives. It compensates for missing modality-specific information within a constrained memory space and aligns with the teacher model. Extensive experiments on IEMOCAP, CMU-MOSI, and CMU-MOSEI demonstrate that PriMD achieves state-of-the-art performance and consistently stronger robustness across a wide range of missing-modality settings, while mitigating the instability caused by holistic feature inference. Our code and project website are available at
\href{https://www.jiaqizhang-sengoku.com/Code/PriMD}{here} and
\href{https://www.jiaqizhang-sengoku.com/PriMD/}{here}, respectively.
\end{abstract}

\section{Introduction}

\begin{figure}[!t]
\centering
\includegraphics[width=0.48\textwidth]{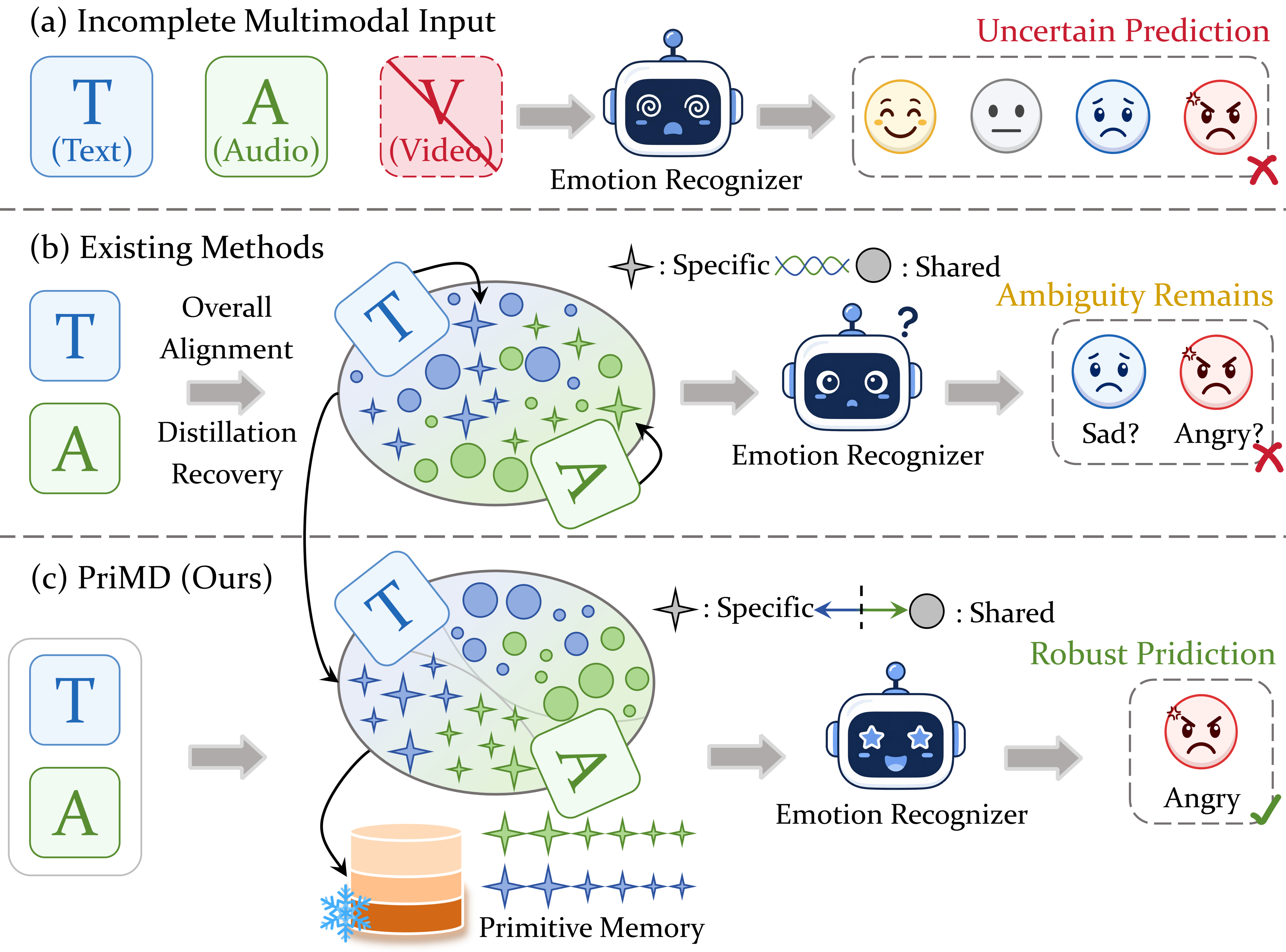}
\caption{
(a) Real-world inputs may suffer from partial modality missingness, leading to uncertain predictions.
(b) Existing holistic generation, alignment, or distillation methods treat missing modalities as a whole. This tends to mix modality-shared semantics with uncertain modality-specific details, leaving the learned representations ambiguous.
(c) Our PriMD first disentangles shared semantics from modality-specific information and performs constrained compensation for the missing modality-specific information. This enables more robust predictions under missing-modality conditions.
}
\label{Figure1}
\end{figure}

Multimodal Emotion Recognition (MER) aims to integrate textual, acoustic, and visual signals to understand human emotional states~\cite{DBLP:journals/inffus/GandhiAPCH23, DBLP:conf/icmi/MorencyMD11}. It has broad applications in ~\cite{DBLP:books/acm/19/KirchnerFK19,lu2026mllms,lu2026worldcoder}, dialogue systems~\cite{DBLP:journals/aiopen/FuGZWY22, DBLP:journals/ai/LiangMZCXZ22}, and social media analysis~\cite{DBLP:journals/pieee/SomandepalliGMK21,DG-SCL}. Compared with unimodal inputs, multimodal inputs provide richer and complementary emotional cues. However, in real-world scenarios, privacy constraints~\cite{DBLP:conf/acl/AguilarRWW19,DBLP:conf/acl/ZhaoLJ20}, sensor failures~\cite{DBLP:conf/acii/VazquezRodriguezLCC23,DBLP:journals/tai/WangCZCYZ23}, and low-quality inputs~\cite{DBLP:journals/tai/WangCZCYZ23, dong2026lance} often cause some modalities to be missing, weakening the robustness of multimodal models.

To address incomplete modalities, existing methods for incomplete multimodal emotion recognition can be broadly grouped into three categories. Modality completion methods recover missing features from available modalities through generators, reconstruction networks, or cross-modal mappings~\cite{DBLP:journals/corr/KingmaW13, DBLP:journals/corr/GoodfellowPMXWOCB14, DBLP:conf/nips/HoJA20, DBLP:conf/iclr/LipmanCBNL23,DPG,PIRP}. However, when the missing modality contains details that are difficult to determine from other modalities, the generated features can be unreliable~\cite{DBLP:journals/inffus/LinH24, DBLP:conf/aaai/YaoYCLQ24, DBLP:conf/cvpr/LinWCLFTYK24,LEADer}. Representation learning methods impose consistency constraints to map different missing-modality combinations into a unified latent space~\cite{DBLP:conf/aaai/PhamLMMP19, DBLP:conf/acl/ZhaoLJ20, DBLP:journals/pami/LianCSLT23,IFCLNet}. This improves model adaptability, but may suppress modality-specific information~\cite{DBLP:conf/mm/XuJL24, DBLP:conf/icml/0001DWC0F024, xu2026lever, dong2026generated}. Knowledge distillation methods use a full-modality teacher model to guide missing-modality student models through feature, relation, or prediction-distribution distillation~\cite{DBLP:conf/aaai/LiYL0WSYWSZ24, DBLP:conf/cvpr/LiYZWWYSKQZ24, DBLP:conf/nips/LiYL0CWWJXHSQKZ24, zhuang2025cmad, liu2026conmem, liu2026dual}. However, most existing distillation strategies focus on transferring teacher knowledge, without distinguishing inferable shared semantics from uncertain modality-specific details. Although these methods bring notable improvements, they still treat missing modalities as a whole during generation, alignment, or distillation. As shown in Figure~\ref{Figure1}, such holistic processing mixes inferable shared semantics with uncertain modality-specific information. This leaves ambiguity in the representations and leads to unstable predictions under missing-modality conditions.

We argue that the above issue arises because information in a missing modality is not equally recoverable. Specifically, the representation of a modality usually contains both cross-modal shared semantics and modality-specific details. The former describes emotional information that is relatively consistent across modalities and can usually be reliably estimated. The latter is tied to the expression form of a specific modality and is therefore more uncertain. Thus, when a missing modality is generated, aligned, or distilled as a whole, recoverable shared information and uncertain modality-specific details are processed together. This makes it difficult to obtain stable representations under complex missing-modality combinations. Although prior studies on full-modality MER have used feature disentanglement to mitigate modality heterogeneity~\cite{DBLP:conf/mm/HazarikaZP20, DBLP:conf/cvpr/LiW023, DBLP:journals/corr/abs-2503-11892,AgentTraces,MAP-Graph,Faulty}, these methods mainly focus on representation learning with full-modality inputs and cannot recover modality-specific information under missing conditions. Based on this observation, we argue that incomplete MER should estimate shared semantics while modeling uncertain modality-specific information in a constrained manner.

To address these issues,  we propose Primitive Memory Distillation (PriMD), a teacher-student framework for structured missing-modality compensation. PriMD does not treat missing-modality recovery as an isolated generation problem. Instead, it integrates recovery into structured representation learning and teacher-student knowledge transfer. In the full-modality teacher model network, we design Shared-Specific Semantic Decoupling (SSSD) to learn and disentangle cross-modal shared semantics and modality-specific representations. Discrete Primitive Memory Construction (DPMC) further discretizes modality-specific representations into learnable semantic primitives and constructs modality-specific primitive memory banks. Under missing-modality conditions, Dynamic Retrieval-Augmented Distillation (DRAD) dynamically retrieves relevant semantic primitives according to the shared semantics of available modalities. It then performs constrained compensation for missing modality-specific information and aligns the student model with the full-modality teacher, leading to stable representations and predictions. The main contributions of this paper are as follows:

\begin{itemize}
\item We revisit incomplete MER from an intra-modal perspective and formulate
it as a unified process of shared-semantic estimation and constrained
compensation of modality-specific information.
\item We design SSSD to disentangle shared and modality-specific features,
and DPMC/DRAD to discretize the modality-specific space into semantic
primitives and retrieve them for compensation under missing modalities.
\item Extensive experiments on three benchmarks show that PriMD outperforms
prior methods under diverse missing-modality settings, with especially large
gains when two modalities are absent.
\end{itemize}

% \begin{itemize}
%     \item We propose PriMD, a framework grounded in the internal information structure of modalities and formulates incomplete MER as a unified process of shared semantic estimation and constrained compensation of modality-specific information.
%     \item PriMD includes SSSD for separating shared and specific features, as well as DPMC and DRAD for discretizing the feature space and addressing missing-modality compensation.
%     \item Extensive experiments show that PriMD mitigates the instability caused by holistic representation and outperforms existing methods under various missing-modality scenarios.
% \end{itemize}

\section{Related Work}
\textbf{Multimodal Learning in MER.} Multimodal Emotion Recognition (MER) aims to learn discriminative emotion representations from heterogeneous signals, such as text, speech, and vision. Early methods mainly focused on modality fusion. TFN~\cite{DBLP:conf/emnlp/ZadehCPCM17} models unimodal and cross-modal interactions through tensor outer products, while LMF~\cite{DBLP:conf/acl/MorencyLZLSL18} uses low-rank factorization to reduce computational cost. As sequential context and modality asynchrony received increasing attention, later methods further introduced memory mechanisms and cross-modal attention. MFN ~\cite{DBLP:conf/aaai/ZadehLMPCM18} introduces multi-view memory units to model cross-modal interactions over time, whereas MulT ~\cite{DBLP:conf/acl/TsaiBLKMS19} adopts directional cross-modal attention to handle unaligned multimodal sequences. To alleviate modality redundancy, noise interference, and semantic inconsistency, recent studies have started to improve model robustness at the representation level. MISR~\cite{DBLP:conf/acl/ZhuHWY0025a} mitigates bias in emotion distributions through invariance learning, while DecAlign~\cite{DBLP:journals/corr/abs-2503-11892} models modality-shared semantics and modality-specific components, preserving both semantic consistency and modality differences. While effective, these methods presuppose full-modality inputs and cannot be directly applied when modalities are absent.

\textbf{Incomplete Multimodal Learning in MER.} Missing modalities are a common issue in real-world multimodal emotion recognition. They are usually caused by sensor failures, noise interference, privacy constraints, or automatic recognition errors. Existing methods can be broadly categorized into three categories: modality completion, robust representation learning, and knowledge distillation. CRA~\cite{DBLP:conf/cvpr/Tran0ZJ17} models modality correlations through cascaded residual autoencoders, while IMDer~\cite{DBLP:conf/nips/WangL023} uses a score-based diffusion model to generate missing features. MPLMM~\cite{DBLP:conf/acl/Guo0Z24} and P-RMF~\cite{DBLP:conf/acl/ZhuHWY0025} enhance missing-information reconstruction through prompt learning and latent proxy modalities, respectively. Representation learning methods avoid explicit generation and instead learn the joint distribution of modalities. MCTN~\cite{DBLP:conf/aaai/PhamLMMP19} learns cross-modal representations through cyclic translation, MMIN~\cite{DBLP:conf/acl/ZhaoLJ20} uses cycle-consistency constraints to compensate for missing information, and GCNet~\cite{DBLP:journals/pami/LianCSLT23} further introduces graph neural networks to model temporal dependencies. In recent years, knowledge distillation has also been adopted to transfer supervision from full-modality teachers models to missing-modality students models~\cite{gou2021knowledge,yang2025survey}. For example, UMDF~\cite{DBLP:conf/aaai/LiYL0WSYWSZ24}, CorrKD~\cite{DBLP:conf/cvpr/LiYZWWYSKQZ24}, and HRLF~\cite{DBLP:conf/nips/LiYL0CWWJXHSQKZ24} improve knowledge transfer from the perspectives of self-distillation, correlation distillation, and hierarchical mutual-information alignment, respectively. However, most existing methods focus on the holistic generation, alignment, and distillation of missing modalities, while paying limited attention to the characterization of modality-specific and shared information. When available modalities cannot reliably infer the missing specific details, such holistic processing may introduce irrelevant or even conflicting features.

\begin{figure*}[!t]
\centering
\includegraphics[width=\textwidth]{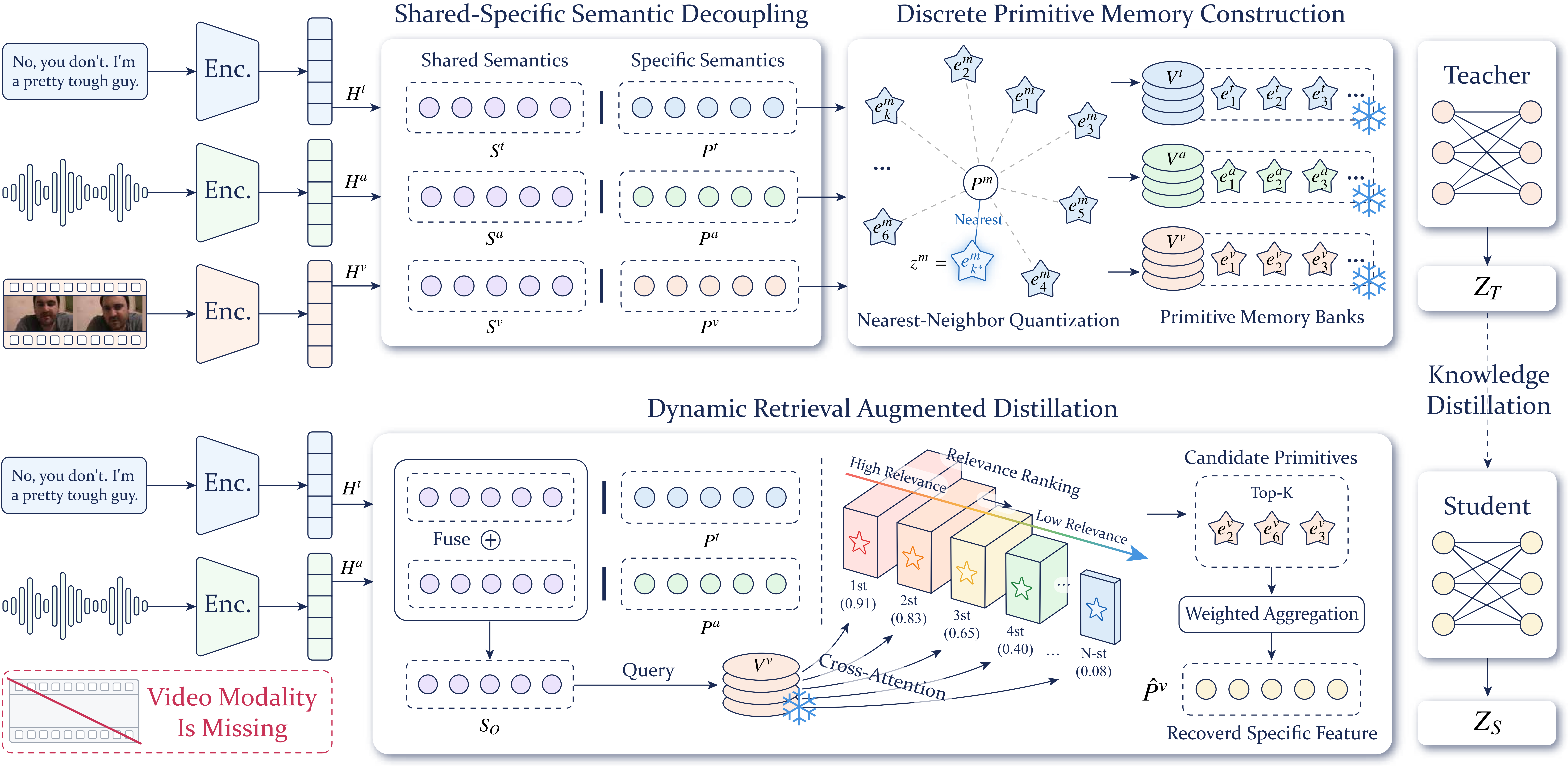}
\caption{\textbf{Overall framework of PriMD.}
The full-modality teacher model first encodes the inputs and disentangles shared semantics from modality-specific information via SSSD. Then, DPMC quantizes the modality-specific representations into discrete semantic primitives and constructs primitive memory banks. In missing-modality scenarios, the student model uses DRAD to dynamically retrieve relevant primitives from the memory banks based on the semantics of available modalities, enabling constrained compensation for the missing information. Finally, the compensated student model representations are aligned with the teacher model through retrieval-augmented distillation.}
\label{Figure2}
\end{figure*}

\section{Methodology}
\textbf{Problem Formulation.} This paper considers a multimodal emotion recognition task with three modalities: textual,
acoustic, and visual signals. Let the full-modality set be \(\Omega=\{t,a,v\}\). For a given sample, its multimodal input is represented as \(X=\{X^m \mid m\in\Omega\}\), where \(X^m=\{x_i^m\}_{i=1}^{T_m}\) denotes the sequential features of modality \(m\). In the incomplete-modality setting, only a subset of modalities is available to the model. We use \(O\subseteq\Omega\) to denote the available modality set and \(M=\Omega\setminus O\) to denote the missing modality set, where \(O\neq\emptyset\), \(O\cup M=\Omega\), and \(O\cap M=\emptyset\). Accordingly, \(X_O=\{X^m \mid m\in O\}\) denotes the available modality input, while \(X_M=\{X^m \mid m\in M\}\) denotes the missing-modality feature set. To represent different missing-modality combinations, we use a modality mask \(c=[c_t,c_a,c_v]\), where \(c_m=1\) indicates that modality \(m\) is available, and \(c_m=0\) indicates that it is missing.

\textbf{Overview.} As shown in Figure~\ref{Figure2}, the proposed PriMD consists of a full-modality teacher model and an incomplete-modality student model. The teacher model first disentangles complete multimodal representations into cross-modal shared semantics and modality-specific information, and then constructs a discrete primitive memory based on the modality-specific information. In the student model, when some modalities are missing, the model dynamically retrieves relevant primitives according to the shared semantics of the available modalities to compensate for the missing modality-specific information. Finally, PriMD aligns the student model with the full-modality teacher model
via retrieval-augmented distillation.
%Finally, PriMD aligns the teacher model with the student model through retrieval-augmented distillation.
%蒸馏是让 student 对齐 teacher，不是反过来

\subsection{Shared-Specific Semantic Decoupling}
Multimodal representations usually contain both cross-modal shared semantics and modality-specific information. Effectively disentangling these two types of information can reduce redundant feature interference in subsequent knowledge transfer and provide a more stable representation basis for missing-modality compensation. To this end, we first use the full-modality teacher network \(F_T\) to encode the input of each modality. Given the complete input \(X^m\), the modality encoder extracts deep features:
\begin{equation}
H^m = \psi_{\mathrm{enc}}^m(X^m) \in \mathbb{R}^{N \times d}, 
\end{equation}
where \(N\) denotes the number of samples in a mini-batch. Then, two independent projection branches map \(H^m\) into the shared semantic representation \(S^m\) and the modality-specific representation \(P^m\):
\begin{align}
S^m &= \psi_s(H^m) \in \mathbb{R}^{N \times d}, \\
P^m &= \psi_p(H^m) \in \mathbb{R}^{N \times d}.
\end{align}

To encourage \(S^m\) to capture cross-modal shared information, we minimize the distance between shared representations from different modalities:
\begin{equation}
L_{\mathrm{Align}}=\frac{2}{N|\Omega|(|\Omega|-1)}\sum_{m\in\Omega}\sum_{n>m}\left\|S^m-S^n\right\|_F^2 .
\end{equation}

% \begin{equation}
% {L}_{\mathrm{Align}}
% =
% \frac{1}{|\Omega|(|\Omega|-1)}
% \sum_{m \in \Omega}
% \sum_{\substack{n \in \Omega \\ n \neq m}}
% \left\| S^m - S^n \right\|_F^2.
% \end{equation}

Meanwhile, to suppress the statistical dependence between the shared semantic representation and the modality-specific representation, and to make \(P^m\) focus more on modality-specific information, we map the features into a reproducing kernel Hilbert space and use the Hilbert-Schmidt Independence Criterion as a constraint:
\begin{equation}
L_{\mathrm{Indep}}
=
\sum_{m \in \Omega}
\frac{1}{(N-1)^2}
\operatorname{Tr}
\left(
K_S^m C_N K_P^m C_N
\right),
\end{equation}
where \(C_N = I_N - \frac{1}{N}\mathbf{1}_N\mathbf{1}_N^{\top}\) denotes the centering matrix, and \(\operatorname{Tr}(\cdot)\) denotes the matrix trace. \(K_S^m\) and \(K_P^m\) are the Gram matrices of \(S^m\) and \(P^m\) under the radial basis function kernel mapping, respectively. \(I_N \in \mathbb{R}^{N \times N}\) is the identity matrix, and \(\mathbf{1}_N \in \mathbb{R}^{N}\) is an all-one vector. After disentanglement, the teacher network concatenates the shared and modality-specific features of all modalities and outputs \(Z_T\) through the classifier \(\Phi_{\mathrm{Cls}}^T\):
\begin{equation}
Z_T =
\Phi_{\mathrm{Cls}}^T
(
\operatorname{Concat}_{m \in \Omega}
\left(S^m, P^m\right)
).
\end{equation}

\subsection{Discrete Primitive Memory Construction}
Continuous high-dimensional feature spaces usually exhibit large distributional variance. When modalities are missing, directly regressing such continuous features is often difficult to optimize and may introduce strong uncertainty. To reduce the difficulty of missing-information compensation and provide more reliable prior representations, we discretize the disentangled modality-specific representations into a finite number of semantic primitives and construct modality-specific primitive memory banks.

For modality \(m\), we define its learnable primitive memory bank as:
\begin{equation}
V^m
=
\left\{
e_1^m, e_2^m, \ldots, e_C^m
\right\}
\in \mathbb{R}^{C \times d},
\end{equation}
where \(C\) denotes the primitive capacity. Given the modality-specific representation \(P^m=\{p_i^m\}_{i=1}^{N}\in\mathbb{R}^{N\times d}\), we quantize each sample representation \(p_i^m\) to the nearest semantic primitive in the memory bank:
\begin{equation}
k_i^*
=
\arg\min_{1 \leq k \leq C}
\left\| p_i^m - e_k^m \right\|_2^2,
\quad
z_i^m = e_{k_i^*}^m,
\end{equation}
where all quantized primitive representations form \(Z^m=\{z_i^m\}_{i=1}^{N}\in\mathbb{R}^{N\times d}\). To make the discrete primitives in the memory bank fit the distribution of modality-specific representations while preventing continuous representations from deviating excessively from their matched primitives during training, we use a vector quantization loss with a commitment constraint:
\begin{equation}
\begin{aligned}
L_{\mathrm{VQ}} =
& \sum_{m \in \Omega} \Big(
    \left\| \operatorname{sg}(P^m) - Z^m \right\|_F^2 \\
& \quad + \beta \left\| P^m - \operatorname{sg}(Z^m) \right\|_F^2
  \Big),
\end{aligned}
\end{equation}
where \(\operatorname{sg}(\cdot)\) denotes the stop-gradient operation, and \(\beta\) is a tunable parameter. After the teacher stage is completed, the modality-specific primitive memory banks are frozen and used as prior retrieval spaces in the student stage.

\subsection{Dynamic Retrieval-Augmented Distillation}
When modalities are missing, the student network lacks complete input cues. Directly aligning it with the full-modality teacher model may introduce incorrect or unstable feature matching. To address this issue, we retrieve from the frozen primitive memory banks to provide constrained feature compensation for missing modalities. Moreover, considering differences in sample difficulty and missing-modality combinations, a fixed number of retrieved primitives is insufficient for complex scenarios~\cite{song2023deterministic}. Therefore, we design a dynamic retrieval-augmented distillation strategy.

For each missing modality \(m \in M\), the student network \(F_S\) first extracts and fuses shared semantics from the available modality set \(O\):
\begin{equation}
S_O
=
\operatorname{Fuse}
\left(
\left\{
S^o \mid o \in O
\right\}
\right)
\in \mathbb{R}^{N \times d}.
\end{equation}
where $\mathrm{Fuse}(\cdot)$ denotes the available-modality shared-semantics fusion operator. Then, using \(S_O\) as the query, we compute the relevance scores between \(S_O\) and the primitives in the memory bank \(V^m\) of the missing modality \(m\):
\begin{equation}
A^m
=
\operatorname{Softmax}
(
\frac{
(S_O W_q)(V^m W_k)^{\top}
}{
\sqrt{d_r}
}
)
\in \mathbb{R}^{N \times C}.
\end{equation}

Here, \(W_q\) and \(W_k\) map the shared semantics and primitives into the same retrieval space, respectively. \(d_r\) is the dimension of the retrieval space, and \(\operatorname{Softmax}(\cdot)\) is applied along the primitive dimension. For the \(b\)-th sample, we sort the primitives in the memory bank in descending order according to its relevance scores \(A_b^m\). The primitive ranked at position \(k\) is denoted as \(\tilde{e}_{b,k}^m\), and its attention weight is denoted as \(\tilde{a}_{b,k}^m\). To dynamically determine the number of retrieved primitives, we use
\(\alpha^m=\operatorname{MLP}(S_O)\in \mathbb{R}^{N \times K_{\max}}\)
to predict retrieval-number logits for the missing modality \(m\), and then generate a differentiable retrieval-number distribution:
\begin{equation}
\pi_{b,r}^m
=
\frac{
\exp
\left(
(\alpha_{b,r}^m+g_{b,r}^m)/\gamma
\right)
}{
\sum_{j=1}^{K_{\max}}
\exp
\left(
(\alpha_{b,j}^m+g_{b,j}^m)/\gamma
\right)
},
\end{equation}
where \(g_{b,r}^m\sim\operatorname{Gumbel}(0,1)\), \(\gamma\) is a hyperparameter controlling the smoothness of the distribution, and \(r\in\{1,\ldots,K_{\max}\}\) denotes a candidate retrieval number. For candidate retrieval number \(r\), we aggregate the top-\(r\) primitives and normalize their attention weights:
\begin{equation}
\phi_{b,r}^m
=
\frac{
\sum_{k=1}^{r}
\tilde{a}_{b,k}^m \tilde{e}_{b,k}^m
}{
\sum_{k=1}^{r}
\tilde{a}_{b,k}^m
+
\epsilon
}.
\end{equation}

Finally, the compensated representation of the missing modality \(m\) for the \(b\)-th sample is obtained as:
\begin{equation}
\hat{P}^m
=
\left(
\sum_{r=1}^{K_{\max}}
\pi_{b,r}^m\phi_{b,r}^m
\right)_{b=1}^{N}
\in \mathbb{R}^{N \times d}.
\end{equation}

After obtaining the compensated representations, the student network concatenates the shared semantics of the available modalities, the modality-specific representations of the available modalities, and compensated missing-modality representations. The concatenated representation is fed into the classifier $\Phi_{\mathrm{Cls}}^S$ to produce the student logits:

\begin{equation}
\begin{aligned}
Z_S &= \Phi_{\mathrm{Cls}}^S
\\ & \Big(
      \operatorname{Concat}\Big(
        S_O, \operatorname*{Concat}_{o \in O} P^o,
        \operatorname*{Concat}_{m \in M} \hat{P}^m
      \Big)
    \Big).
\end{aligned}
\end{equation}

To prevent the model from always selecting the maximum retrieval number, we introduce a nonlinear cost penalty for retrieval:
\begin{equation}
{L}_{\mathrm{Cost}}
=
\frac{1}{N}
\sum_{b=1}^{N}
\sum_{m\in M}
\sum_{r=1}^{K_{\max}}
\pi_{b,r}^m
(
\exp(\frac{r}{K_{\max}}) - 1
).
\end{equation}

%Table 1 中 HARDY-MER 两列数字完全相同

%MOSEI 行：A-only = 74.82 / 74.11，A+V = 74.82 / 74.11。两个不同设置数值一模一样，几乎肯定是复制错误。请回原文核对

% Table 1 与 Table 7 的列顺序不一致
% Table 1 双模态顺序：\mc{1}{1}{0}(A+T), \mc{1}{0}{1}(A+V), \mc{0}{1}{1}(T+V)
% Table 7 双模态顺序：\mc{1}{0}{1}(A+V), \mc{1}{1}{0}(A+T), \mc{0}{1}{1}

\begin{table*}[!t]
\caption{Performance comparison with SOTA methods on two datasets under various missing-modality scenarios. "Avg." indicates the mean performance across all settings. The \colorbox{best1}{best} and \colorbox{best2}{suboptimal} results are highlighted. A, T, and V denote the acoustic, textual, and visual modalities, respectively. Gray shading indicates the presence of a modality, while gray dashed lines indicate its absence.}
\label{Table1}
\centering
\definecolor{best1}{HTML}{D7EAFB}
\definecolor{best2}{HTML}{E6F4EA}
\renewcommand{\arraystretch}{1.2}
\setlength{\tabcolsep}{4pt}
\setlength{\fboxsep}{1pt}
\setlength{\aboverulesep}{0pt}
\setlength{\belowrulesep}{0pt}
\resizebox{\linewidth}{!}{%
\begin{tabular}{cl cccccccccccccc}
\toprule
\multirow{2}{*}{Dataset} & \multirow{2}{*}{Method}
& \multicolumn{2}{c}{\mc{1}{0}{0}} & \multicolumn{2}{c}{\mc{0}{1}{0}} & \multicolumn{2}{c}{\mc{0}{0}{1}}
& \multicolumn{2}{c}{\mc{1}{1}{0}} & \multicolumn{2}{c}{\mc{1}{0}{1}} & \multicolumn{2}{c}{\mc{0}{1}{1}}
& \multicolumn{2}{c}{Avg.} \\
\cmidrule(lr){3-4} \cmidrule(lr){5-6} \cmidrule(lr){7-8} \cmidrule(lr){9-10} \cmidrule(lr){11-12} \cmidrule(lr){13-14} \cmidrule(lr){15-16}
& & WA(\%) & UA(\%) & WA(\%) & UA(\%) & WA(\%) & UA(\%)
& WA(\%) & UA(\%) & WA(\%) & UA(\%) & WA(\%) & UA(\%)
& WA(\%) & UA(\%) \\
\midrule
\multirow{8}{*}{{\shortstack{IEMOCAP\\Four Class}}}
& CPMNet~\cite{DBLP:journals/pami/ZhangCHZFH22}
& 46.85 & 51.72 & 45.63 & 45.32 & 44.95 & 44.49 & 34.81 & 36.23 & 48.67 & 49.33 & 45.62 & 46.57 & 44.42 & 45.61 \\
& MMIN~\cite{DBLP:conf/acl/ZhaoLJ20}
& 56.58 & 59.00 & 66.57 & 68.02 & 52.52 & 50.60 & 72.94 & 71.14 & 63.99 & 63.43 & 71.67 & 68.61 & 64.05 & 63.47 \\
& GCNet~\cite{DBLP:journals/pami/LianCSLT23}
& 65.58 & 68.76 & 72.33 & 70.42 & 57.96 & 52.54 & 77.02 & 76.87 & 67.40 & 65.64 & 75.63 & 73.62 & 69.32 & 67.98 \\
& CIF-MMIN~\cite{DBLP:journals/taffco/LiuZLSL24}
& 57.53 & 60.06 & 67.22 & 68.99 & 53.46 & 51.56 & 74.19 & 72.59 & 64.99 & 63.53 & 72.40 & 69.91 & 64.97 & 64.44 \\
& MoMKE~\cite{DBLP:conf/mm/XuJL24}
& 69.53 & 70.21 & 77.30 & 77.66 & 56.80 & 52.03 & 79.03 & 79.88 & 68.57 & 66.22 & 75.55 & 74.18 & 71.13 & 70.03 \\
& HARDY-MER~\cite{DBLP:conf/mm/0008Z0YF25}
& \colorbox{best2}{72.65} & \colorbox{best2}{73.87} & \colorbox{best2}{82.49} & \colorbox{best2}{82.69} & \colorbox{best2}{63.19} & \colorbox{best2}{60.54} & \colorbox{best2}{81.67} & \colorbox{best2}{82.43} & \colorbox{best2}{74.19} & \colorbox{best2}{74.50} & \colorbox{best2}{79.18} & \colorbox{best2}{78.51} & \colorbox{best2}{75.56} & \colorbox{best2}{75.42} \\
\cdashline{2-16}
& PriMD (Ours)
& \colorbox{best1}{73.82} & \colorbox{best1}{74.46} & \colorbox{best1}{84.13} & \colorbox{best1}{84.09} & \colorbox{best1}{64.77} & \colorbox{best1}{62.81} & \colorbox{best1}{83.65} & \colorbox{best1}{84.83} & \colorbox{best1}{75.06} & \colorbox{best1}{74.84} & \colorbox{best1}{83.17} & \colorbox{best1}{82.66} & \colorbox{best1}{77.43} & \colorbox{best1}{77.28} \\
\midrule
\multirow{8}{*}{{\shortstack{IEMOCAP\\Six Class}}}
& CPMNet~\cite{DBLP:journals/pami/ZhangCHZFH22}
& 29.47 & 29.80 & 32.44 & 34.95 & 26.20 & 24.95 & 33.49 & 33.94 & 26.92 & 25.46 & 31.34 & 30.43 & 29.98 & 29.92 \\
& MMIN~\cite{DBLP:conf/acl/ZhaoLJ20}
& 44.08 & 42.96 & 42.17 & 38.55 & 35.74 & 30.65 & 51.95 & 48.31 & 41.92 & 38.15 & 47.49 & 40.63 & 43.89 & 39.88 \\
& GCNet~\cite{DBLP:journals/pami/LianCSLT23}
& 49.95 & 46.45 & 56.48 & 55.62 & 39.78 & 34.97 & 58.24 & 57.25 & 47.57 & 43.31 & 57.43 & 54.66 & 51.58 & 48.71 \\
& CIF-MMIN~\cite{DBLP:journals/taffco/LiuZLSL24}
& 44.96 & 43.56 & 43.40 & 39.71 & 36.11 & 31.35 & 52.43 & 49.20 & 42.54 & 39.22 & 48.88 & 44.91 & 44.72 & 41.33 \\
& MoMKE~\cite{DBLP:conf/mm/XuJL24}
& 50.51 & 47.38 & 61.09 & 60.19 & 39.07 & 34.51 & 63.18 & 61.94 & 48.65 & 44.08 & 59.92 & 57.55 & 53.74 & 50.94 \\
& HARDY-MER~\cite{DBLP:conf/mm/0008Z0YF25}
& \colorbox{best2}{51.58} & \colorbox{best2}{49.14} & \colorbox{best2}{65.89} & \colorbox{best2}{61.95} & \colorbox{best2}{43.02} & \colorbox{best2}{36.49} & \colorbox{best2}{65.18} & \colorbox{best2}{62.98} & \colorbox{best2}{52.91} & \colorbox{best2}{47.45} & \colorbox{best2}{61.66} & \colorbox{best2}{57.86} & \colorbox{best2}{56.71} & \colorbox{best2}{52.65} \\
\cdashline{2-16}
& PriMD (Ours)
& \colorbox{best1}{53.62} & \colorbox{best1}{51.43} & \colorbox{best1}{66.21} & \colorbox{best1}{62.17} & \colorbox{best1}{45.35} & \colorbox{best1}{39.86} & \colorbox{best1}{65.78} & \colorbox{best1}{63.66} & \colorbox{best1}{54.17} & \colorbox{best1}{49.82} & \colorbox{best1}{63.65} & \colorbox{best1}{60.77} & \colorbox{best1}{58.13} & \colorbox{best1}{54.62} \\
\midrule

& & ACC(\%) & F1(\%) & ACC(\%) & F1(\%) & ACC(\%) & F1(\%)
& ACC(\%) & F1(\%) & ACC(\%) & F1(\%) & ACC(\%) & F1(\%)
& ACC(\%) & F1(\%) \\
\midrule

\multirow{9}{*}{{\shortstack{CMU\\MOSEI}}}
& CPMNet~\cite{DBLP:journals/pami/ZhangCHZFH22}
& 65.71 & 65.18 & 72.87 & 72.44 & 61.23 & 61.73 & 72.65 & 72.24 & 61.56 & 61.99 & 66.29 & 66.84 & 66.72 & 66.74 \\
& MMIN~\cite{DBLP:conf/acl/ZhaoLJ20}
& 58.90 & 59.50 & 82.20 & 82.40 & 59.30 & 60.01 & 83.70 & 83.30 & 63.55 & 61.91 & 81.75 & 81.42 & 71.57 & 71.42 \\
& GCNet~\cite{DBLP:journals/pami/LianCSLT23}
& 72.04 & 70.34 & 84.26 & 84.17 & 68.08 & 67.25 & 85.10 & 85.10 & 71.49 & 69.96 & 84.74 & 84.54 & 77.62 & 76.89 \\
& CIF-MMIN~\cite{DBLP:journals/taffco/LiuZLSL24}
& 63.87 & 64.60 & 83.53 & 83.04 & 61.96 & 62.66 & 84.01 & 83.47 & 64.68 & 62.08 & 82.50 & 81.94 & 73.43 & 72.97 \\
& MoMKE~\cite{DBLP:conf/mm/XuJL24}
& 72.56 & 71.03 & 86.10 & 86.03 & 64.50 & 63.46 & \colorbox{best2}{86.32} & \colorbox{best2}{86.29} & 72.37 & 72.07 & \colorbox{best1}{86.90} & \colorbox{best2}{86.91} & 78.13 & 77.63 \\
& CMAD~\cite{zhuang2025cmad}
& 63.00 & 60.80 & 86.10 & 86.00 & 65.70 & 64.40 & 86.30 & 86.20 & 65.60 & 64.80 & 86.40 & 86.40 & 75.52 & 74.77 \\
& HARDY-MER~\cite{DBLP:conf/mm/0008Z0YF25}
& \colorbox{best1}{74.82} & \colorbox{best2}{74.11} & \colorbox{best2}{87.20} & \colorbox{best2}{87.13} & \colorbox{best1}{69.35} & \colorbox{best1}{67.50} & 85.42 & 85.01 & \colorbox{best1}{74.82} & \colorbox{best1}{74.11} & 85.72 & 85.39 & \colorbox{best2}{79.56} & \colorbox{best2}{78.88} \\
\cdashline{2-16}
& PriMD (Ours)
& \colorbox{best2}{74.61} & \colorbox{best1}{74.25} & \colorbox{best1}{87.36} & \colorbox{best1}{87.24} & \colorbox{best2}{68.96} & \colorbox{best2}{67.36} & \colorbox{best1}{87.26} & \colorbox{best1}{87.69} & \colorbox{best2}{73.85} & \colorbox{best2}{73.61} & \colorbox{best2}{86.81} & \colorbox{best1}{87.27} & \colorbox{best1}{79.81} & \colorbox{best1}{79.57} \\
\bottomrule
\end{tabular}%
}
\end{table*}

Finally, we use the output \(Z_T\) of the frozen full-modality teacher model as soft labels to distill knowledge into the student model:
\begin{equation}
\begin{aligned}
L_{\mathrm{KD}}
&= \tau_{\mathrm{KD}}^2 \cdot D_{\mathrm{KL}}\Bigl(
\operatorname{Softmax}(Z_T / \tau_{\mathrm{KD}}) \\
&\quad \Big\| \operatorname{Softmax}(Z_S / \tau_{\mathrm{KD}})
\Bigr),
\end{aligned}
\end{equation}
where \(\tau_{\mathrm{KD}}\) is the distillation temperature.

\subsection{Joint Optimization}
We train PriMD with a two-stage optimization strategy. In the teacher stage, we use full-modality data to optimize the full-modality teacher network \(F_T\) and the modality-specific primitive memory banks \(V\). After training, the teacher network and memory banks are frozen and used as the knowledge source and retrieval space for the student stage. In the student stage, we train the student network \(F_S\) with missing-modality masks, enabling it to perform dynamic retrieval, feature compensation, and knowledge alignment under incomplete-modality conditions. The optimization objectives of the two stages are defined as:
\begin{equation}
\begin{aligned}
L_{\mathrm{Tea}} =
& L_{\mathrm{Task}}^T
+ \lambda_{\mathrm{Align}} L_{\mathrm{Align}} \\
& + \lambda_{\mathrm{Indep}} L_{\mathrm{Indep}}
+ \lambda_{\mathrm{VQ}} L_{\mathrm{VQ}},
\end{aligned}
\end{equation}
\begin{equation}
{L}_{\mathrm{Stu}}
=
{L}_{\mathrm{Task}}^S
+
{L}_{\mathrm{KD}}
+
{L}_{\mathrm{Cost}}.
\end{equation}
where $\lambda_{\mathrm{Align}}$, $\lambda_{\mathrm{Indep}}$, and $\lambda_{\mathrm{VQ}}$ weight the corresponding losses.

\section{Experiments}
\subsection{Experiment Setup}
\textbf{Datasets.} To evaluate the effectiveness of our proposed method, we conduct extensive experiments on three datasets: IEMOCAP~\cite{DBLP:journals/lre/BussoBLKMKCLN08}, CMU-MOSI~\cite{DBLP:journals/corr/ZadehZPM16}, and CMU-MOSEI~\cite{DBLP:conf/acl/MorencyCPLZ18}.

\textbf{Evaluation Metrics.} 
For IEMOCAP, we report Weighted Accuracy (WA) and Unweighted Accuracy (UA). 
For CMU-MOSI and CMU-MOSEI, we adopt task-specific metrics, including Accuracy (ACC), F1-score, and Mean Absolute Error (MAE), following the corresponding experimental settings.

Details of the datasets and evaluation metrics are shown in Appendix~\ref{A}.

\textbf{Implementation Details.} Detailed experimental setup and parameters are provided in Appendix~\ref{B}.

\subsection{Comparison with SOTA Methods}
\textbf{Comparison under inter-modal missingness.} To evaluate the robustness of PriMD in incomplete multimodal scenarios, we quantitatively compare it with SOTA methods under different missing-modality settings, as shown in Table~\ref{Table1}. PriMD achieves the best average performance across all three experimental settings in terms of the average over the six missing-modality combinations. Specifically, on the four-class IEMOCAP task, PriMD achieves 77.43\% WA and 77.28\% UA, outperforming the second-best method, HARDY-MER, by 1.87\% and 1.86\%, respectively. On the more challenging six-class IEMOCAP task, PriMD achieves 58.13\% WA and 54.62\% UA, improving over HARDY-MER by 1.42\% and 1.97\%, respectively. On CMU-MOSEI, PriMD obtains an average ACC/F1 of 79.81\%/79.57\%, also surpassing existing methods. Overall, these results indicate that holistic reconstruction or direct alignment of missing modalities is susceptible to uncertain modality-specific information. In contrast, PriMD performs constrained compensation for missing-specific information through shared-semantics-guided primitive retrieval, thereby more effectively mitigating representation shifts caused by missing modalities.

\begin{figure*}[!t]
\centering
\includegraphics[width=\textwidth]{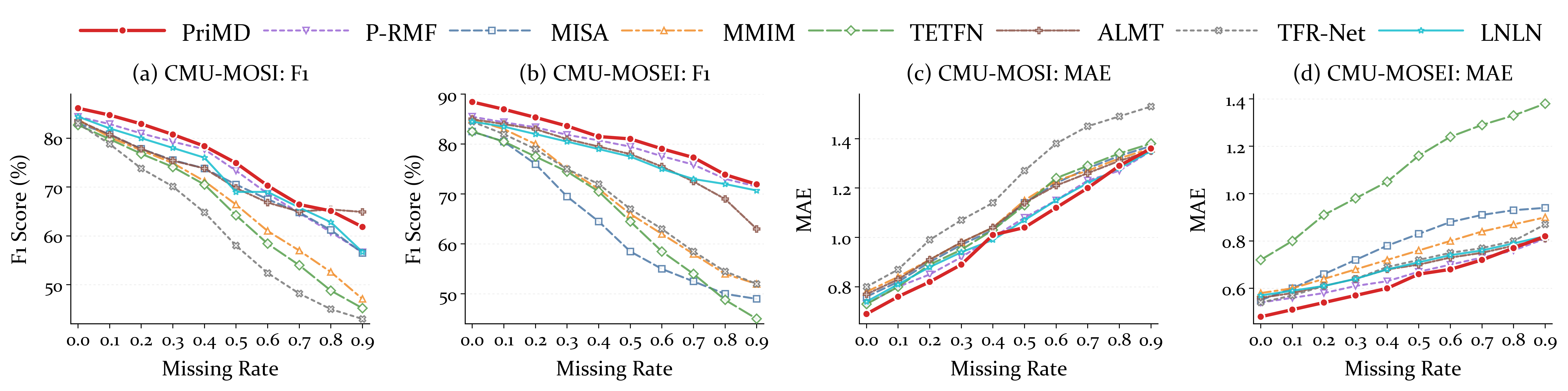} 
\caption{Performance curves of different methods under varying missing rates on the CMU-MOSI and CMU-MOSEI datasets. Subfigures (a)-(b) and (c)-(d) show F1 and MAE results on MOSI and MOSEI, respectively.}
\label{Figure3}
\end{figure*}

\textbf{Comparison for Intra-modal Missingness:} We further evaluate the robustness of PriMD under intra-modal missingness on CMU-MOSI and CMU-MOSEI. Following previous works~\cite{DBLP:conf/nips/ZhangWY24, DBLP:conf/acl/ZhuHWY0025}, we gradually increase the missing rate from 0.0 to 0.9. As shown in Figure~\ref{Figure3}, the performance of most methods decreases as the missing rate increases, while PriMD maintains higher F1 scores and lower MAE under most missing-rate settings. In contrast, P-RMF mitigates the information deficiency via proxy modalities, but its compensation still operates on holistic representations and therefore making it vulnerable to uncertain details when intra-modal information is continuously lost. For LNLN, its F1 score improves under some settings as the missing rate increases, whereas its MAE continues to increase. This suggests that its classification results may be affected by data bias, making the model more likely to predict high-frequency classes rather than learn stable affective representations. PriMD shows more stable performance across different missing rates, indicating that it can capture shared semantics from the remaining modalities and reduce the representation shift caused by intra-modal information loss.

\subsection{Ablation Study}
To verify the effectiveness of the core modules in PriMD, we conducted ablation experiments. The results are shown in Table~\ref{Table2}.

\begin{table*}[!t]
\caption{Results of the ablation experiments under the Four-Class IEMOCAP. The \colorbox{best1}{best} results are highlighted.}
\label{Table2}
\centering

\definecolor{best1}{HTML}{D7EAFB}
\definecolor{best2}{HTML}{E6F4EA} 
\renewcommand{\arraystretch}{1.2} 
\setlength{\tabcolsep}{4pt}
\setlength{\fboxsep}{1pt}

\setlength{\aboverulesep}{0pt}
\setlength{\belowrulesep}{0pt}

\resizebox{\linewidth}{!}{%
\begin{tabular}{cl cccccccccccccc}
\toprule

\multirow{2}{*}{Dataset} & \multirow{2}{*}{Modules}
& \multicolumn{2}{c}{\mc{1}{0}{0}} & \multicolumn{2}{c}{\mc{0}{1}{0}} & \multicolumn{2}{c}{\mc{0}{0}{1}}
& \multicolumn{2}{c}{\mc{1}{1}{0}} & \multicolumn{2}{c}{\mc{1}{0}{1}} & \multicolumn{2}{c}{\mc{0}{1}{1}}
& \multicolumn{2}{c}{Avg.} \\
\cmidrule(lr){3-4} \cmidrule(lr){5-6} \cmidrule(lr){7-8} \cmidrule(lr){9-10} \cmidrule(lr){11-12} \cmidrule(lr){13-14} \cmidrule(lr){15-16}

& & WA(\%) & UA(\%) & WA(\%) & UA(\%) & WA(\%) & UA(\%)
& WA(\%) & UA(\%) & WA(\%) & UA(\%) & WA(\%) & UA(\%)
& WA(\%) & UA(\%) \\
\midrule

\multirow{4.5}{*}{{\shortstack{IEMOCAP\\Four Class}}}
& w/o SSSD
& 70.15 & 70.82 & 81.25 & 81.02 & 61.34 & 59.45 & 80.77 & 81.12 & 72.41 & 72.05 & 80.22 & 79.85 & 74.36 & 74.05 \\
& w/o DPMC
& 72.41 & 73.15 & 82.88 & 82.91 & 63.12 & 61.05 & 81.95 & 82.74 & 73.88 & 73.61 & 81.45 & 81.12 & 75.95 & 75.76 \\
& w/o DRAD
& 69.88 & 70.02 & 79.45 & 79.12 & 58.76 & 57.11 & 78.33 & 79.65 & 69.52 & 69.18 & 77.94 & 77.53 & 72.31 & 72.10 \\
\cdashline{2-16}
& PriMD (Ours)
& \colorbox{best1}{73.82} & \colorbox{best1}{74.46} & \colorbox{best1}{84.13} & \colorbox{best1}{84.09} & \colorbox{best1}{64.77} & \colorbox{best1}{62.81} & \colorbox{best1}{83.65} & \colorbox{best1}{84.83} & \colorbox{best1}{75.06} & \colorbox{best1}{74.84} & \colorbox{best1}{83.17} & \colorbox{best1}{82.66} & \colorbox{best1}{77.43} & \colorbox{best1}{77.28} \\
\bottomrule
\end{tabular}%
}
\end{table*}

\textbf{1) w/o SSSD:} To evaluate the role of shared-specific semantic disentanglement, we replace the two projection branches with a single projection, so that each modality yields one entangled representation, without separating cross-modal shared semantics from modality-specific information. The average performance of this variant drops to 74.36\% WA and 74.05\% UA. The results show that directly mixing these two types of information for subsequent memory construction and distillation makes the model vulnerable to redundant or unstable modality-specific information, thereby weakening its representation robustness under missing-modality scenarios.

\textbf{2) w/o DPMC:} To analyze the contribution of discrete primitive memory construction, we removed DPMC. Compared with the full model, this variant decreases the average WA/UA to 75.95\%/75.76\%, respectively. This indicates that modality-specific information in the continuous feature space is uncertain, and directly performing generation or alignment makes it difficult to obtain stable representations. DPMC provides a more constrained compensation space for missing specific information through primitive memory.

\textbf{3) w/o DRAD:} To verify the contribution of dynamic retrieval-augmented distillation, we remove the entire DRAD module. In this variant, the student model uses only the shared and modality-specific representations of the
available modalities. The results show that this setting leads
to the most severe performance degradation, with average results of only
72.31\% WA and 72.10\% UA. This indicates that representations from the
available modalities alone are insufficient to adapt to different
missing-modality combinations.

\subsubsection{Decomposition and Compensation}

As shown in Table~\ref{TableDecompComp}(a), we constructed a capacity-matched variant, Holistic-PriMD, to evaluate the role of shared-specific semantic disentanglement. This variant constructs the memory banks directly from holistic modality representations and queries them using the fused representation of the available modalities. Compared with Holistic-PriMD, PriMD improved WA/UA from 75.02\%/74.81\% to 77.43\%/77.28\%, while reducing cross-combination JS and teacher--student KL from 0.091/0.121 to 0.066/0.081. These results indicate that shared-specific semantic disentanglement improves recognition performance and reduces prediction shifts across missing-modality combinations and teacher--student discrepancies under matched model capacity.

As shown in Table~\ref{TableDecompComp}(b), we further separated the contribution of semantic disentanglement from that of the compensation mechanism by comparing SSSD+MLP, SSSD+CVAE, continuous compensation, and discrete primitive retrieval under the same SSSD framework. PriMD achieved the highest WA/UA of 77.43\%/77.28\% and the lowest JS/KL of 0.066/0.081. Moreover, its normalized within-class dispersion was 0.64, compared with 1.00 for continuous compensation. These results indicate that discrete primitive retrieval provides an independent benefit under the same disentanglement framework and is associated with a more compact within-class distribution.

\begin{table*}[!t]
\centering
\caption{Analysis of shared-specific decomposition and feature compensation on Four-Class IEMOCAP. Disp. denotes normalized within-class dispersion. The \colorbox{best1}{best} results are highlighted.}
\label{TableDecompComp}
\renewcommand{\arraystretch}{1.0}
\setlength{\tabcolsep}{3pt}

% 保持较小的字号
\footnotesize 

\begin{tabular*}{\linewidth}{@{\extracolsep{\fill}} l c c c c @{}}
\toprule
\textit{(a) Shared-specific decomposition} & WA (\%) & UA (\%) & \makecell{Cross-combination JS $\downarrow$} & \makecell{Teacher--student KL $\downarrow$} \\
\midrule
Holistic-PriMD & $75.02 \pm 0.58$ & $74.81 \pm 0.61$ & $0.091 \pm 0.004$ & $0.121 \pm 0.006$ \\
PriMD (Ours) & \colorbox{best1}{$77.43 \pm 0.43$} & \colorbox{best1}{$77.28 \pm 0.47$} & \colorbox{best1}{$0.066 \pm 0.003$} & \colorbox{best1}{$0.081 \pm 0.004$} \\
\bottomrule
\end{tabular*}

\vspace{12pt}
\begin{tabular*}{\linewidth}{@{\extracolsep{\fill}} l c c c c c @{}}
\toprule
\textit{(b) Feature compensation} & WA (\%) & UA (\%) & Disp. $\downarrow$ & \makecell{Cross-combination JS $\downarrow$} & \makecell{Teacher--student KL $\downarrow$} \\
\midrule
SSSD+MLP & $75.98 \pm 0.52$ & $75.81 \pm 0.56$ & -- & $0.089 \pm 0.005$ & $0.118 \pm 0.007$ \\
SSSD+CVAE & $76.27 \pm 0.49$ & $76.08 \pm 0.53$ & -- & $0.082 \pm 0.004$ & $0.109 \pm 0.006$ \\
Continuous compensation & $76.34 \pm 0.48$ & $75.83 \pm 0.52$ & 1.00 & $0.079 \pm 0.004$ & $0.106 \pm 0.006$ \\
PriMD (Ours) & \colorbox{best1}{$77.43 \pm 0.43$} & \colorbox{best1}{$77.28 \pm 0.47$} & \colorbox{best1}{0.64} & \colorbox{best1}{$0.066 \pm 0.003$} & \colorbox{best1}{$0.081 \pm 0.004$} \\
\bottomrule
\end{tabular*}
\end{table*}

\begin{table*}[!t]
\centering
\caption{Computational efficiency comparison.}
\label{TableEfficiency}

% 添加与上一个表格统一的字号
\footnotesize 

\begin{tabular*}{\textwidth}{@{\extracolsep{\fill}} lcccc @{}}
\toprule
Method & Params. & Train (s) & Memory (MiB) & Infer. (ms) \\
\midrule
HARDY-MER    & 11.52M                  & 90.981                   & 1839.4                  & \colorbox{best1}{26.829} \\
PriMD (Ours) & \colorbox{best1}{2.34M} & \colorbox{best1}{44.205} & \colorbox{best1}{327.4} & 29.013 \\
\bottomrule
\end{tabular*}
\end{table*}

\begin{table}[!t]
\centering
\renewcommand{\arraystretch}{1.0}
\caption{Ablation results of SSSD components.}
\resizebox{0.48\textwidth}{!}{%
\begin{tabular}{c c c c c c}
\toprule
\multicolumn{2}{c}{\makecell[c]{SSSD}}
&
\multicolumn{2}{c}{\makecell[c]{IEMOCAP \\
Four Class}}
& \multicolumn{2}{c}{\makecell[c]{CMU \\
MOSEI}} \\
\cmidrule(lr){1-2} \cmidrule(lr){3-4} \cmidrule(lr){5-6}
$L_{\mathrm{Align}}$ & $L_{Indep}$ &  WA(\%) & UA(\%) &  ACC(\%)  & F1(\%) \\
\midrule
\checkmark&  & 75.39 & 75.87 & 77.82 & 78.06 \\
& \checkmark & 75.96 & 76.24 & 77.58 & 77.64  \\
\midrule
\rowcolor{gray!5} \checkmark & \checkmark &  \colorbox{best1}{77.43}  &  \colorbox{best1}{77.28} & \colorbox{best1}{79.81} &  \colorbox{best1}{79.57} \\
\bottomrule
\end{tabular}%
}
\label{Table3}
\end{table}

\subsubsection{SSSD Components}

\textbf{4) Ablation Study on SSSD:} To further verify the role of different constraints in SSSD, we separately remove the shared semantic alignment loss $L_{\mathrm{Align}}$ and the independence constraint $L_{Indep}$. As shown in Table~\ref{Table3}, using only $L_{\mathrm{Align}}$ or $L_{Indep}$ leads to lower performance than the full setting. This indicates that $L_{\mathrm{Align}}$ helps improve the consistency of shared semantics across modalities, while $L_{Indep}$ reduces redundant correlations between shared and modality-specific representations. By combining them, the model achieves more effective disentanglement.

\subsection{Computational Efficiency}

As shown in Table~\ref{TableEfficiency}, PriMD reduces the parameter count, training time, and peak CUDA memory by 79.7\%, 51.4\%, and 82.2\%, respectively, compared with HARDY-MER, while increasing inference latency by 8.1\%.

\subsection{Hyperparameter Sensitivity}

\begin{figure}[!t]
    \centering
    \includegraphics[width=1.0\columnwidth]{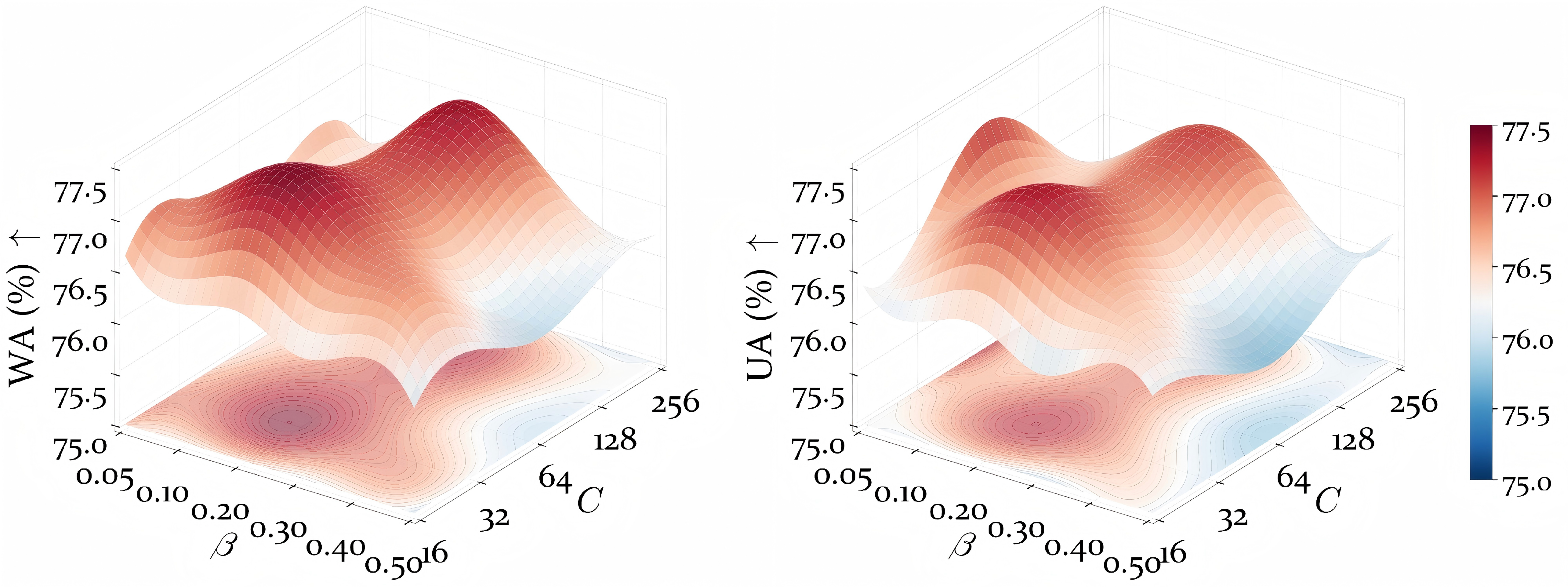}
    \caption{Parameter sensitivity on $C$ and $\beta$ of PriMD.}
    \label{Figure4}
\end{figure}

We evaluate the sensitivity of PriMD to the primitive capacity $C$ and $\beta$ in DPMC. As shown in Figure~\ref{Figure4}, PriMD remains stable across a wide range of settings. A smaller $C$ limits the diversity of specific primitives, while an excessively large $C$ may introduce redundant primitives and reduce retrieval reliability. Similarly, a smaller $\beta$ provides insufficient constraints between modality-specific features and discrete primitives, whereas a larger $\beta$ over-constrains the representation space. These results show that PriMD benefits from a primitive space with balanced constraints for missing-modality compensation.

\subsection{Maximum Retrieval Size}
\begin{figure}[htbp]
    \centering
    \includegraphics[width=1.0\columnwidth]{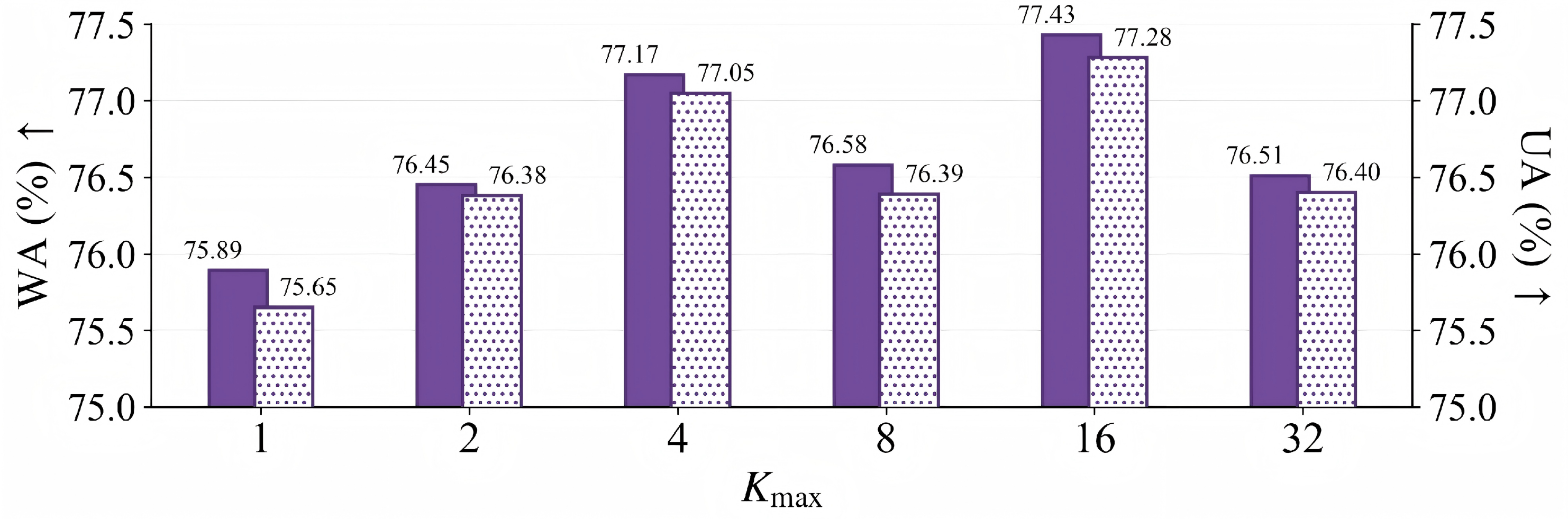}
    \caption{Performance under different $K_{max}$.}
    \label{Figure5}
\end{figure}

We further analyze the effect of $K_{\max}$, which controls the maximum number of primitives retrieved by DRAD. As shown in Figure~\ref{Figure5}, a smaller $K_{\max}$ restricts the student model to very limited modality-specific evidence. Increasing $K_{\max}$ allows the model to consider more complementary primitives and improves the robustness of compensation. However, an excessively large $K_{\max}$ may introduce weakly relevant primitives and dilute the reconstructed representation.

\section{Conclusion}
In this paper, we proposed PriMD, a primitive memory distillation framework for incomplete multimodal emotion recognition. Unlike holistic modality-inference methods, PriMD first decomposes multimodal representations into cross-modal shared semantics and modality-specific information, and then discretizes the latter into modality-specific primitive memories. Under missing-modality conditions, the student model leverages the shared semantics of available modalities to retrieve relevant primitives. This enables constrained compensation for uncertain modality-specific cues and effective distillation from the full-modality teacher model. Extensive experiments showed that PriMD achieves robust performance under both inter-modal and intra-modal missingness. It effectively alleviates the instability caused by holistic feature inference and provides a new perspective for robust incomplete multimodal learning.

\section*{Acknowledgments}
We thank the anonymous reviewers for their constructive feedback.

\section*{Limitations}
Although PriMD achieves good robustness under various incomplete multimodal settings, several limitations remain. First, PriMD relies on a full-modality teacher model to learn shared-specific representations and construct modality-specific primitive memory banks. When full-modality training samples are scarce, biased, or costly to collect, the learned primitives may not fully cover the modality-specific information in the target domain. To reduce this dependence, future work may combine cross-modal self-supervised pretraining or semi-supervised learning to make better use of incomplete samples. Confidence-weighted primitive updates may also allow high-confidence incomplete samples to expand the memory banks. Domain adaptation could further reduce the distribution gap between the training and target domains. Second, our current experiments mainly use sample-level fixed missing masks to simulate inter-modal and intra-modal missingness. In real applications, the set of available modalities may change over time. For example, audio or video sensors may fail temporarily, and input quality may gradually decrease during continuous interaction. Such online missingness can be represented by a time-varying modality mask. PriMD does not yet explicitly model the dependence between missing states at different time steps. Future work will explore time-aware retrieval, modality-quality gating, historical primitive caching, and online memory updates. These methods may allow the model to continuously adjust its compensation strategy according to real-time modality availability and quality, improving reliability and generalization in practical deployment.

\bibliography{Reference}

\clearpage

\appendix

\section{Datasets and Evaluation Metrics}
\label{A}
%Table~\ref{IEMOCAP} 求和：四分类 1085+1023+1151+1031+1241 = 5,531；六分类 = 7,433
To verify the effectiveness of the proposed method, we conduct extensive
experiments on three mainstream multimodal emotion recognition datasets.
As shown in Table~\ref{IEMOCAP}, the IEMOCAP dataset
\cite{DBLP:journals/lre/BussoBLKMKCLN08} contains five dyadic sessions
of improvised or scripted dialogues involving ten actors, with two actors
in each session. Following existing research settings
\cite{DBLP:conf/mm/0008Z0YF25,DBLP:journals/taffco/LiuZLSL24},
we evaluate the model on two utterance-level multi-class emotion
classification tasks. The four-class setting contains 5,531 utterances
from happy (with excited merged into happy), sad, neutral, and angry
\cite{DBLP:conf/mm/0008Z0YF25,DBLP:conf/mm/XuJL24,
DBLP:journals/taffco/LiuZLSL24,DBLP:conf/icassp/ZuoLZGL23,
DBLP:conf/acl/ZhaoLJ20}. The six-class setting contains 7,433 utterances
from happy, sad, neutral, angry, excited, and frustrated
\cite{DBLP:conf/mm/0008Z0YF25,DBLP:conf/aaai/Mai0X20,
DBLP:journals/pami/LianCSLT23,DBLP:conf/aaai/MajumderPHMGC19}.
For both IEMOCAP tasks, we adopt WA and UA as evaluation metrics.

As shown in Table~\ref{CMU-MOSI-CMU-MOSEI}, the CMU-MOSI dataset~\cite{DBLP:journals/corr/ZadehZPM16} contains 2,199 opinionated monologue video clips collected from YouTube, split into 1,284 training samples, 229 validation samples, and 686 test samples. Each utterance-level clip is annotated with a continuous emotion intensity score in $[-3, 3]$, ranging from highly negative to highly positive. As an extension of CMU-MOSI, the CMU-MOSEI dataset~\cite{DBLP:conf/acl/MorencyCPLZ18} is a larger and more challenging corpus. It contains 22,856 video clips covering diverse topics, unconstrained environments, and different speakers, with 16,326 training samples, 1,871 validation samples, and 4,659 test samples. CMU-MOSEI uses the same continuous emotion intensity annotations in $[-3, 3]$. For CMU-MOSI and CMU-MOSEI, the evaluation metrics are selected according to the specific task setting. For binary emotion classification, we follow the standard non-zero binary classification protocol~\cite{DBLP:conf/mm/XuJL24}, where samples with scores greater than 0 are labeled as positive, samples with scores less than 0 are labeled as negative, and zero-score samples are excluded from binary evaluation. In the inter-modal missingness comparison and the noise robustness experiments, we report ACC and F1 for binary emotion classification. Specifically, CMU-MOSEI is evaluated with ACC and F1 in the main inter-modal missingness comparison, while both CMU-MOSI and CMU-MOSEI are evaluated with ACC and F1 in the noise robustness analysis. For the intra-modal missingness experiments on CMU-MOSI and CMU-MOSEI, following previous work~\cite{DBLP:conf/acl/ZhuHWY0025}, we report weighted F1 for binary emotion classification and MAE for emotion intensity prediction. MAE is computed between the predicted and gold emotion scores in $[-3, 3]$, where lower values indicate better regression performance.

\begin{table}[!t]
\centering
\resizebox{\columnwidth}{!}{
\begin{tabular}{c c c c}
\toprule
\multicolumn{2}{c}{Dataset} & \# Conversations & \# Utterances \\
\midrule
\multirow{5}{*}{\begin{tabular}[c]{@{}c@{}}IEMOCAP\\ Four-Class\end{tabular}} & Session1 & 28 & 1085 \\
 & Session2 & 30 & 1023 \\
 & Session3 & 32 & 1151 \\
 & Session4 & 30 & 1031 \\
 & Session5 & 31 & 1241 \\
\midrule
\multirow{5}{*}{\begin{tabular}[c]{@{}c@{}}IEMOCAP\\ Six-Class\end{tabular}} & Session1 & 28 & 1373 \\
 & Session2 & 30 & 1356 \\
 & Session3 & 32 & 1569 \\
 & Session4 & 30 & 1512 \\
 & Session5 & 31 & 1623 \\
\bottomrule
\end{tabular}
}
\caption{Statistical information on IEMOCAP.}
\label{IEMOCAP}
\end{table}

\begin{table}[!t]
\centering
\resizebox{\columnwidth}{!}{
\begin{tabular}{l c c ccc ccc}
\toprule
\multirow{2.5}{*}{Dataset} & \multirow{2.5}{*}{Speaker} & \multirow{2.5}{*}{Video Clip} & \multicolumn{3}{c}{\# Conversations} & \multicolumn{3}{c}{\# Utterances} \\
\cmidrule(lr){4-6} \cmidrule(lr){7-9}
 & & & train & val & test & train & val & test \\
\midrule
CMU-MOSI  & 89   & 2199  & 52   & 10  & 31  & 1284  & 229  & 686 \\
CMU-MOSEI & 1000 & 22856 & 2249 & 300 & 676 & 16326 & 1871 & 4659 \\
\bottomrule
\end{tabular}
}
\caption{Statistical information on CMU-MOSI and CMU-MOSEI.}
\label{CMU-MOSI-CMU-MOSEI}
\end{table}

\section{Implementation Details}
\label{B}

\subsection{Comparison Methods} 
For the inter-modal missingness setting, we compare PriMD with CPMNet~\cite{DBLP:journals/pami/ZhangCHZFH22}, MMIN~\cite{DBLP:conf/acl/ZhaoLJ20}, GCNet~\cite{DBLP:journals/pami/LianCSLT23}, CIF-MMIN~\cite{DBLP:journals/taffco/LiuZLSL24}, MoMKE~\cite{DBLP:conf/mm/XuJL24}, CMAD~\cite{zhuang2025cmad}, and HARDY-MER~\cite{DBLP:conf/mm/0008Z0YF25}.

For the intra-modal missingness setting, we compare PriMD with P-RMF~\cite{DBLP:conf/acl/ZhuHWY0025}, MISA~\cite{DBLP:conf/mm/HazarikaZP20}, MMIM~\cite{DBLP:conf/emnlp/HanCP21}, TETFN~\cite{DBLP:journals/pr/WangGTLHL23}, ALMT~\cite{DBLP:conf/emnlp/ZhangWYL0Y23}, TFR-Net~\cite{DBLP:conf/mm/YuanLXY21}, and LNLN~\cite{DBLP:conf/nips/ZhangWY24}. For these methods, we likewise do not re-implement or re-evaluate them. Instead, we mainly refer to the publicly reported results in LNLN~\cite{DBLP:conf/nips/ZhangWY24} and P-RMF~\cite{DBLP:conf/acl/ZhuHWY0025}. 

Across the aforementioned experimental settings, we ensure that the pre-extracted multimodal features, data splits, and evaluation metrics are exactly the same as those used by the compared methods. To maintain consistency with prior work~\cite{DBLP:conf/mm/XuJL24,DBLP:conf/acl/ZhuHWY0025,DBLP:conf/mm/0008Z0YF25}, we adopt the same protocol for fair comparison.

\subsection{PriMD Implementation}
To ensure the fairness of the experimental results, for the inter-modal missingness setting, we adopt the publicly available pre-extracted features provided by prior studies~\cite{DBLP:conf/mmm/LuoXL23,DBLP:conf/nips/WangL023,DBLP:conf/acl/ZhaoLJ20,DBLP:conf/icassp/ZuoLZGL23}, and evaluate the models under six fixed combinations of missing modalities. For the intra-modal missingness setting, we follow existing work~\cite{DBLP:conf/nips/ZhangWY24,DBLP:conf/acl/ZhuHWY0025}. We use their available pre-extracted features and gradually increase the missing rate from 0.0 to 0.9 with an increment of 0.1 to simulate missing data. Specifically, when the missing rate is 0.5, 50\% of the information in each modality of the test data is randomly masked. PriMD is trained using PyTorch 2.7.0 with CUDA 12.8 on an NVIDIA RTX 5090 GPU with 32~GB of memory. For the IEMOCAP dataset, we perform five-fold cross-validation using the Leave-One-Session-Out strategy. This is a subject-independent evaluation protocol, since each session contains two unique actors who do not appear in any other session.
For the CMU-MOSI and CMU-MOSEI datasets, we repeat each experiment five times and report the averaged results. The proposed framework is optimized with the AdamW optimizer. The batch size is set to 32, the initial learning rate is set to $1 \times 10^{-4}$, and the weight decay is set to $1 \times 10^{-5}$. The maximum number of training epochs is set to 100. Other detailed parameter settings of our method are reported in Table~\ref{Hyperparameters}.

\begin{table*}[!t]
\centering
\caption{Hyperparameter settings of PriMD.}
\label{Hyperparameters}
\renewcommand{\arraystretch}{1.0}
\setlength{\tabcolsep}{12pt}
\begin{tabular}{@{} l ccc @{}}
\toprule
{Hyperparameter} & {IEMOCAP} & {CMU-MOSI} & {CMU-MOSEI} \\
\cmidrule(r){1-1} \cmidrule(lr){2-2} \cmidrule(lr){3-3} \cmidrule(l){4-4}
Optimizer & AdamW & AdamW & AdamW \\
Learning rate & $1\times10^{-4}$ & $1\times10^{-4}$ & $1\times10^{-4}$ \\
Weight decay & $1\times10^{-5}$ & $1\times10^{-5}$ & $1\times10^{-5}$ \\
Batch size & 32 & 32 & 32 \\
Epoch & 100 & 100 & 100 \\
$d$ & 128 & 128 & 128 \\
$d_r$ & 64 & 64 & 64 \\
Dropout rate & 0.3 & 0.3 & 0.3 \\
$C$ & 64 & 64 & 64 \\
$\beta$ & 0.20 & 0.20 & 0.20 \\
$K_{\max}$ & 16 & 16 & 16 \\
$\gamma$ & 0.5 & 0.5 & 0.5 \\
$\tau_{\mathrm{KD}}$ & 4.0 & 4.0 & 4.0 \\
$\lambda_{\mathrm{Align}}$ & 0.1 & 0.1 & 0.1 \\
$\lambda_{\mathrm{Indep}}$ & 0.01 & 0.01 & 0.01 \\
$\lambda_{\mathrm{VQ}}$ & 0.1 & 0.1 & 0.1 \\
\bottomrule
\end{tabular}
\end{table*}

\section{Extended Main Experiments}

\label{app:extended-main}

\subsection{Noise Robustness}

To further evaluate the robustness of PriMD under input perturbations, following prior work~\cite{DBLP:conf/cvpr/Gao00S0S24}, we add controlled noise with varying intensities to the multimodal features during training, validation, and testing. As shown in Table~\ref{Table7}, we compare our method with MoMKE~\cite{DBLP:conf/mm/XuJL24} and DiCMoR~\cite{DBLP:conf/iccv/WangCL23} across multiple experimental settings. The performance of all methods decreases as the noise intensity increases. This indicates that noise corrupts the original modality representations and further increases the difficulty of incomplete multimodal emotion recognition. Compared with existing methods, PriMD maintains more stable average performance on CMU-MOSI, CMU-MOSEI, and the IEMOCAP Four-Class setting. For example, under strong noise with $\sigma=20$, PriMD achieves average ACC/F1 scores of 53.67\%/53.91\% on MOSI and 61.93\%/61.68\% on MOSEI, as well as average WA/UA scores of 34.50\%/34.14\% on IEMOCAP, outperforming the compared methods overall. These results show that, when noise interference and modality missingness coexist, PriMD can reduce representation shifts caused by unreliable features and produce more robust predictions.

\begin{table*}[!t]
\caption{Performance comparison on MOSI, MOSEI, and IEMOCAP datasets under varying noise levels ($\sigma$). The \colorbox{best1}{best} results are highlighted. A, T, and V denote the acoustic, textual, and visual modalities, respectively. Gray shading indicates the presence of a modality, while gray dashed lines indicate its absence.}
\label{Table7}
\centering
\definecolor{best1}{HTML}{D7EAFB}
\definecolor{best2}{HTML}{E6F4EA}
\renewcommand{\arraystretch}{1.2}
\setlength{\tabcolsep}{4pt}
\setlength{\fboxsep}{1pt}
\setlength{\aboverulesep}{0pt}
\setlength{\belowrulesep}{0pt}
\resizebox{\linewidth}{!}{%
\begin{tabular}{clc cccccccccccccc}
\toprule
\multirow{2}{*}{Dataset} & \multirow{2}{*}{Method} & \multirow{2}{*}{$\sigma$}
& \multicolumn{2}{c}{\mc{1}{0}{0}} & \multicolumn{2}{c}{\mc{0}{1}{0}} & \multicolumn{2}{c}{\mc{0}{0}{1}}
& \multicolumn{2}{c}{\mc{1}{1}{0}} & \multicolumn{2}{c}{\mc{1}{0}{1}} & \multicolumn{2}{c}{\mc{0}{1}{1}}
& \multicolumn{2}{c}{Avg.} \\
\cmidrule(lr){4-5} \cmidrule(lr){6-7} \cmidrule(lr){8-9} \cmidrule(lr){10-11} \cmidrule(lr){12-13} \cmidrule(lr){14-15} \cmidrule(lr){16-17}
& & & ACC(\%) & F1(\%) & ACC(\%) & F1(\%) & ACC(\%) & F1(\%)
& ACC(\%) & F1(\%) & ACC(\%) & F1(\%) & ACC(\%) & F1(\%) & ACC(\%) & F1(\%) \\
\midrule
\multirow{9}{*}{MOSI}
& DiCMoR & 5
& \colorbox{best1}{53.24} & 47.90 & 70.58 & 70.51 & 56.02 & 49.02 & 72.87 & 72.99 & 55.09 & 54.72 & 69.97 & 69.75 & 62.96 & 60.82 \\
& MoMKE & 5
& 52.90 & 52.25 & 58.69 & 58.90 & 52.13 & 51.83 & 59.49 & 59.41 & 52.29 & 51.79 & 57.77 & 57.93 & 55.55 & 55.35 \\
& PriMD & 5
& 53.13 & \colorbox{best1}{52.84} & \colorbox{best1}{73.06} & \colorbox{best1}{73.81} & \colorbox{best1}{57.64} & \colorbox{best1}{56.27} & \colorbox{best1}{73.89} & \colorbox{best1}{73.93} & \colorbox{best1}{56.03} & \colorbox{best1}{57.12} & \colorbox{best1}{74.56} & \colorbox{best1}{75.06} & \colorbox{best1}{64.72} & \colorbox{best1}{64.84} \\
\cdashline{2-17}
& DiCMoR & 10
& 50.46 & 45.91 & \colorbox{best1}{59.72} & 57.88 & 51.85 & 47.91 & 61.57 & 58.04 & 54.63 & 50.62 & \colorbox{best1}{63.43} & 59.81 & 56.94 & 53.36 \\
& MoMKE & 10
& 51.68 & 51.69 & 54.42 & 54.69 & 53.96 & 54.18 & 54.27 & 54.33 & 53.81 & 54.05 & 54.57 & 54.78 & 53.79 & 53.95 \\
& PriMD & 10
& \colorbox{best1}{52.36} & \colorbox{best1}{52.47} & 59.44 & \colorbox{best1}{60.02} & \colorbox{best1}{54.69} & \colorbox{best1}{55.34} & \colorbox{best1}{62.15} & \colorbox{best1}{62.43} & \colorbox{best1}{55.04} & \colorbox{best1}{55.27} & 62.19 & \colorbox{best1}{63.08} & \colorbox{best1}{57.65} & \colorbox{best1}{58.10} \\
\cdashline{2-17}
& DiCMoR & 20
& 48.32 & 44.69 & 50.00 & 47.67 & 47.87 & 45.36 & 50.46 & 46.80 & 42.68 & 38.71 & 50.15 & 48.34 & 48.25 & 45.26 \\
& MoMKE & 20
& 52.44 & 52.61 & 54.27 & 54.51 & 53.35 & 53.64 & \colorbox{best1}{52.13} & 52.42 & 51.22 & 51.36 & 53.35 & 53.60 & 52.79 & 53.02 \\
& PriMD & 20
& \colorbox{best1}{53.54} & \colorbox{best1}{53.46} & \colorbox{best1}{55.27} & \colorbox{best1}{55.61} & \colorbox{best1}{54.16} & \colorbox{best1}{54.27} & 52.09 & \colorbox{best1}{52.73} & \colorbox{best1}{52.69} & \colorbox{best1}{52.45} & \colorbox{best1}{54.28} & \colorbox{best1}{54.94} & \colorbox{best1}{53.67} & \colorbox{best1}{53.91} \\
\midrule
\multirow{6}{*}{MOSEI}
& MoMKE & 5
& 60.48 & 55.91 & 77.74 & 77.67 & 63.87 & 60.83 & 77.66 & 77.10 & 63.04 & 59.63 & 77.11 & 76.94 & 69.98 & 68.01 \\
& PriMD & 5
& \colorbox{best1}{61.83} & \colorbox{best1}{56.97} & \colorbox{best1}{79.17} & \colorbox{best1}{78.85} & \colorbox{best1}{64.52} & \colorbox{best1}{64.73} & \colorbox{best1}{79.15} & \colorbox{best1}{79.52} & \colorbox{best1}{66.08} & \colorbox{best1}{66.47} & \colorbox{best1}{79.34} & \colorbox{best1}{79.66} & \colorbox{best1}{71.68} & \colorbox{best1}{71.03} \\
\cdashline{2-17}
& MoMKE & 10
& 60.37 & 55.01 & 62.25 & 60.66 & 61.89 & 55.25 & 60.48 & 59.90 & \colorbox{best1}{60.76} & 56.46 & 61.28 & 61.44 & 61.17 & 58.12 \\
& PriMD & 10
& \colorbox{best1}{61.13} & \colorbox{best1}{60.45} & \colorbox{best1}{64.25} & \colorbox{best1}{65.03} & \colorbox{best1}{63.22} & \colorbox{best1}{62.36} & \colorbox{best1}{64.25} & \colorbox{best1}{65.07} & 60.54 & \colorbox{best1}{61.83} & \colorbox{best1}{67.24} & \colorbox{best1}{67.51} & \colorbox{best1}{63.44} & \colorbox{best1}{63.71} \\
\cdashline{2-17}
& MoMKE & 20
& 59.74 & 55.50 & 60.13 & 55.45 & 57.65 & 55.00 & 58.67 & 56.52 & 58.06 & 54.85 & 60.26 & 54.95 & 59.09 & 55.38 \\
& PriMD & 20
& \colorbox{best1}{61.15} & \colorbox{best1}{60.24} & \colorbox{best1}{63.27} & \colorbox{best1}{63.44} & \colorbox{best1}{59.68} & \colorbox{best1}{59.73} & \colorbox{best1}{61.28} & \colorbox{best1}{61.94} & \colorbox{best1}{60.41} & \colorbox{best1}{59.86} & \colorbox{best1}{65.77} & \colorbox{best1}{64.85} & \colorbox{best1}{61.93} & \colorbox{best1}{61.68} \\
\midrule
& &
& WA(\%) & UA(\%) & WA(\%) & UA(\%) & WA(\%) & UA(\%)
& WA(\%) & UA(\%) & WA(\%) & UA(\%) & WA(\%) & UA(\%) & WA(\%) & UA(\%) \\
\midrule
\multirow{6}{*}{\shortstack{IEMOCAP\\Four Class}}
& MoMKE & 5
& \colorbox{best1}{32.75} & 29.01 & 48.99 & 49.48 & 42.83 & 37.65 & 50.28 & 50.84 & 42.57 & 37.75 & 52.45 & 51.92 & 44.98 & 42.78 \\
& PriMD & 5
& 31.04 & \colorbox{best1}{30.59} & \colorbox{best1}{55.83} & \colorbox{best1}{55.96} & \colorbox{best1}{44.21} & \colorbox{best1}{41.85} & \colorbox{best1}{53.85} & \colorbox{best1}{55.62} & \colorbox{best1}{48.11} & \colorbox{best1}{40.25} & \colorbox{best1}{59.63} & \colorbox{best1}{60.81} & \colorbox{best1}{48.78} & \colorbox{best1}{47.51} \\
\cdashline{2-17}
& MoMKE & 10
& 31.47 & 27.67 & 32.71 & 29.46 & 39.08 & 31.17 & 32.15 & 28.78 & 36.83 & 31.77 & 34.36 & 31.32 & 34.43 & 30.03 \\
& PriMD & 10
& \colorbox{best1}{31.85} & \colorbox{best1}{30.54} & \colorbox{best1}{36.77} & \colorbox{best1}{39.82} & \colorbox{best1}{42.01} & \colorbox{best1}{40.53} & \colorbox{best1}{48.71} & \colorbox{best1}{49.22} & \colorbox{best1}{40.26} & \colorbox{best1}{39.14} & \colorbox{best1}{44.36} & \colorbox{best1}{44.12} & \colorbox{best1}{40.66} & \colorbox{best1}{40.56} \\
\cdashline{2-17}
& MoMKE & 20
& 31.11 & 28.27 & 31.19 & 28.03 & 31.37 & 28.35 & 30.69 & 28.10 & 31.16 & 28.44 & 31.28 & 28.34 & 31.13 & 28.26 \\
& PriMD & 20
& \colorbox{best1}{31.40} & \colorbox{best1}{30.03} & \colorbox{best1}{33.64} & \colorbox{best1}{32.29} & \colorbox{best1}{34.55} & \colorbox{best1}{35.23} & \colorbox{best1}{35.11} & \colorbox{best1}{34.85} & \colorbox{best1}{34.71} & \colorbox{best1}{35.27} & \colorbox{best1}{37.60} & \colorbox{best1}{37.18} & \colorbox{best1}{34.50} & \colorbox{best1}{34.14} \\
\bottomrule
\end{tabular}%
}
\end{table*}

\section{Extended Ablation Study}

\label{app:extended-ablation}

We further supplement our analysis with comprehensive ablation studies on the core module and its internal mechanisms.

\subsection{DPMC Components}

In the ablation study of DPMC, we design four variants to analyze its effects. w/o DPMC removes the entire DPMC module and no longer uses the primitive memory bank. w/o Discrete Primitives removes the discretization process and directly uses continuous modality-specific representations for compensation. Shared Memory Bank allows the three modalities to share a single memory bank, instead of constructing separate modality-specific memory banks. Random Memory Bank uses a randomly initialized memory bank to verify the necessity of learnable primitives. As shown in Table~\ref{DPMC}, the full DPMC achieves the best performance under all missing-modality combinations, with average WA/UA scores of 77.43\%/77.28\%. Compared with w/o DPMC, the full model improves WA and UA by 1.48 and 1.52 percentage points, respectively, demonstrating the effectiveness of primitive memory for missing-modality compensation. The full model also maintains stable advantages over w/o Discrete Primitives and Shared Memory Bank, indicating that discrete primitive modeling and modality-specific memory banks help preserve modality-specific information. The performance of Random Memory Bank is overall lower than that of directly removing the DPMC module, suggesting that a randomly constructed memory bank introduces ineffective primitives, which instead interfere with missing-modality compensation and degrade model performance.

\begin{table*}[!t]
\caption{Ablation results for DPMC components. The \colorbox{best1}{best} results are highlighted.}
\label{DPMC}
\centering

\definecolor{best1}{HTML}{D7EAFB}
\definecolor{best2}{HTML}{E6F4EA} 
\renewcommand{\arraystretch}{1.2} 
\setlength{\tabcolsep}{4pt}
\setlength{\fboxsep}{1pt}

\setlength{\aboverulesep}{0pt}
\setlength{\belowrulesep}{0pt}

\resizebox{\linewidth}{!}{%
\begin{tabular}{cl cccccccccccccc}
\toprule

\multirow{2}{*}{Dataset} & \multirow{2}{*}{Modules}
& \multicolumn{2}{c}{\mc{1}{0}{0}} & \multicolumn{2}{c}{\mc{0}{1}{0}} & \multicolumn{2}{c}{\mc{0}{0}{1}}
& \multicolumn{2}{c}{\mc{1}{1}{0}} & \multicolumn{2}{c}{\mc{1}{0}{1}} & \multicolumn{2}{c}{\mc{0}{1}{1}}
& \multicolumn{2}{c}{Avg.} \\
\cmidrule(lr){3-4} \cmidrule(lr){5-6} \cmidrule(lr){7-8} \cmidrule(lr){9-10} \cmidrule(lr){11-12} \cmidrule(lr){13-14} \cmidrule(lr){15-16}

& & WA(\%) & UA(\%) & WA(\%) & UA(\%) & WA(\%) & UA(\%)
& WA(\%) & UA(\%) & WA(\%) & UA(\%) & WA(\%) & UA(\%)
& WA(\%) & UA(\%) \\
\midrule

\multirow{5.5}{*}{{\shortstack{IEMOCAP\\Four Class}}}
& w/o DPMC
& 72.41 & 73.15 & 82.88 & 82.91 & 63.12 & 61.05 & 81.95 & 82.74 & 73.88 & 73.61 & 81.45 & 81.12 & 75.95 & 75.76 \\
& w/o Discrete Primitives
& 72.66 & 72.93 & 83.54 & 83.72 & 63.41 & 61.06 & 82.61 & 81.82 & 73.89 & 73.74 & 81.94 & 81.73 & 76.34 & 75.83 \\
& Shared Memory Bank
& 72.59 & 73.84 & 83.45 & 83.37 & 63.62 & 61.68 & 82.49 & 83.06 & 74.27 & 73.94 & 82.09 & 81.86 & 76.42 & 76.29 \\
& Random Memory Bank
& 68.45 & 69.62 & 80.43 & 80.46 & 61.35 & 60.13 & 80.47 & 80.96 & 71.88 & 71.65 & 80.15 & 79.84 & 73.79 & 73.78 \\
\cdashline{2-16}
& DPMC
& \colorbox{best1}{73.82} & \colorbox{best1}{74.46} & \colorbox{best1}{84.13} & \colorbox{best1}{84.09} & \colorbox{best1}{64.77} & \colorbox{best1}{62.81} & \colorbox{best1}{83.65} & \colorbox{best1}{84.83} & \colorbox{best1}{75.06} & \colorbox{best1}{74.84} & \colorbox{best1}{83.17} & \colorbox{best1}{82.66} & \colorbox{best1}{77.43} & \colorbox{best1}{77.28} \\
\bottomrule
\end{tabular}%
}
\end{table*}

\subsection{DRAD Retrieval Analysis}

\subsubsection{Fixed Retrieval Size}

To validate the effectiveness of the dynamic retrieval strategy in DRAD, we compare it with Top-$K$ strategies using different fixed retrieval sizes. The fixed Top-$K$ methods use the same number of primitives for all samples, with $K=1,2,4,8,16$. In contrast, the dynamic retrieval method adaptively selects the primitives to retrieve according to sample characteristics, under the maximum retrieval budget of $K_{\max}=16$. As shown in Table~\ref{K}, the performance of fixed Top-$K$ first increases and then decreases as $K$ grows. The best fixed retrieval result is achieved when $K=4$, with WA/UA of $76.86\%/76.65\%$. When $K$ is too small, the number of retrieved primitives is insufficient, making it difficult to adequately compensate for missing-modality information. When $K$ is too large, the additionally introduced primitives may contain redundant or weakly relevant information, which degrades representation quality~\cite{lu2025deepresearch,qian2026relevant,qian2026toward}. By contrast, dynamic retrieval achieves the best performance, with $77.43\%$ WA and $77.28\%$ UA. It improves over the best fixed Top-$K$ setting by $0.57$ and $0.63$ percentage points, respectively. This suggests that adaptively determining the retrieval size for different samples is more effective for missing-modality compensation.

\begin{table}[!t]
\centering
\caption{Effect of fixed Top-$K$ and dynamic retrieval on Four-Class IEMOCAP. The \colorbox{best1}{best} results are highlighted.}
\label{K}

\definecolor{best1}{HTML}{D7EAFB}
\renewcommand{\arraystretch}{1.0}
\setlength{\fboxsep}{1pt}
\setlength{\aboverulesep}{0pt}
\setlength{\belowrulesep}{0pt}
\begin{tabular*}{\linewidth}{@{\extracolsep{\fill}} l cc @{}}
\toprule
{Method} & {WA (\%)} & {UA (\%)} \\
\midrule
\multicolumn{3}{@{}l}{\textit{Fixed Top-K}} \\
\hspace{1em} $K=1$  & 75.84 & 75.62 \\
\hspace{1em} $K=2$  & 76.31 & 76.08 \\
\hspace{1em} $K=4$  & 76.86 & 76.65 \\
\hspace{1em} $K=8$  & 76.58 & 76.41 \\
\hspace{1em} $K=16$ & 76.12 & 75.94 \\
\midrule
\multicolumn{3}{@{}l}{\textit{Dynamic Retrieval}} \\
\hspace{1em} $K_{\max}=16$ & \colorbox{best1}{\textbf{77.43}} & \colorbox{best1}{\textbf{77.28}} \\
\bottomrule
\end{tabular*}
\end{table}

\subsubsection{Retrieval Cost Functions}

As shown in Table~\ref{TableRetrieval}, dynamic retrieval without a cost constraint improves WA/UA to 77.07\%/76.91\%, but uses 10.6 primitives on average. The linear cost reduces the expected retrieval size to 5.9 and yields 77.31\%/77.16\% WA/UA. The exponential cost further reduces the expected size to 5.2 and achieves the highest 77.43\%/77.28\% WA/UA. These results indicate that the exponential cost provides the best balance between recognition performance and primitive usage under the current setting.

\begin{table*}[!t]
\centering
\caption{Comparison of retrieval-cost functions on Four-Class IEMOCAP. The \colorbox{best1}{best} results are highlighted.}
\label{TableRetrieval}
\renewcommand{\arraystretch}{1.1}
\begin{tabular*}{\linewidth}{@{\extracolsep{\fill}} l c c c c c @{}}
\toprule
Retrieval strategy & WA (\%) & UA (\%) & Expected $r$ & Params & Inference (ms) \\
\midrule
Fixed $K=4$ & $76.86 \pm 0.45$ & $76.65 \pm 0.48$ & 4.0 & 2.327M & 28.660 \\
Dynamic, no cost & $77.07 \pm 0.46$ & $76.91 \pm 0.49$ & 10.6 & -- & -- \\
Dynamic, linear cost & $77.31 \pm 0.44$ & $77.16 \pm 0.48$ & 5.9 & -- & -- \\
Dynamic, exponential cost & \colorbox{best1}{$77.43 \pm 0.43$} & \colorbox{best1}{$77.28 \pm 0.47$} & 5.2 & 2.336M & 29.013 \\
\bottomrule
\end{tabular*}
\end{table*}

\subsection{Shared Alignment Objectives}

As shown in Table~\ref{TableAlignment}, we compared cosine, InfoNCE, mutual-information, and $L_2$ alignment while keeping the remaining model architecture and training settings unchanged. Under the current naturally paired multimodal setting, $L_2$ alignment achieved the highest WA/UA of 77.43\%/77.28\%, slightly exceeding the 77.16\%/76.98\% obtained by cosine alignment, while InfoNCE and mutual-information alignment yielded lower performance. These results support the use of $L_2$ alignment in the current setting, but do not imply its universal superiority across datasets or pairing schemes.

\begin{table}[!t]
\centering
\caption{Comparison of shared-alignment objectives on Four-Class IEMOCAP. The \colorbox{best1}{best} results are highlighted.}
\label{TableAlignment}
\renewcommand{\arraystretch}{1.1}
\setlength{\tabcolsep}{3pt}
\begin{tabular*}{\linewidth}{@{\extracolsep{\fill}} l c c @{}}
\toprule
Alignment & WA (\%) & UA (\%) \\
\midrule
Cosine & $77.16 \pm 0.45$ & $76.98 \pm 0.49$ \\
InfoNCE & $76.88 \pm 0.50$ & $76.71 \pm 0.53$ \\
Mutual information & $76.74 \pm 0.52$ & $76.56 \pm 0.55$ \\
$L_2$ & \colorbox{best1}{$77.43 \pm 0.43$} & \colorbox{best1}{$77.28 \pm 0.47$} \\
\bottomrule
\end{tabular*}
\end{table}

\begin{figure}[!t]
    \centering
    \includegraphics[width=0.9\columnwidth]{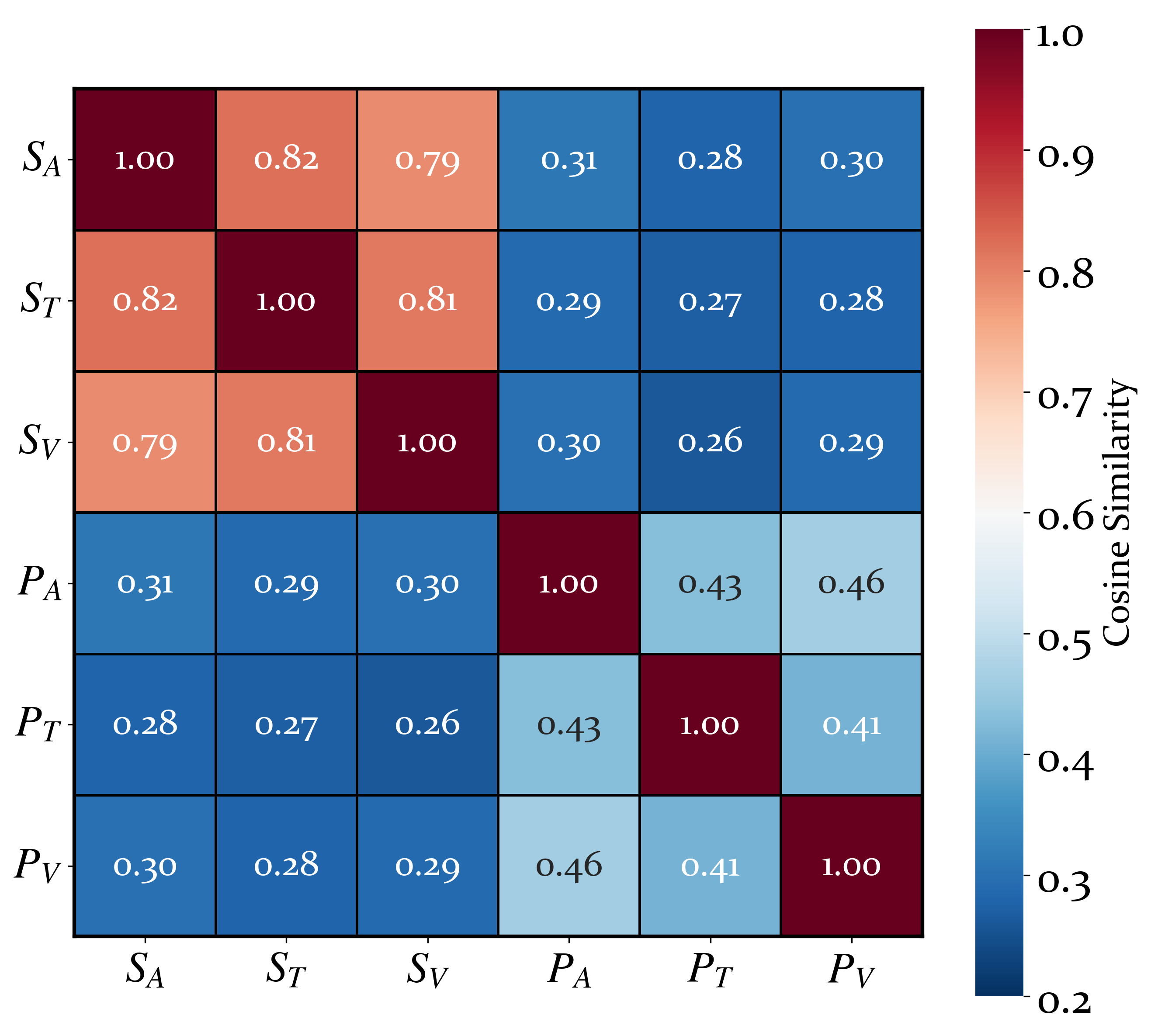}
    \caption{Visualization of shared-specific representation relations on Four-Class IEMOCAP.}
    \label{Figure7}
\end{figure}

\begin{figure*}[!t]

\centering
\includegraphics[width=\textwidth]{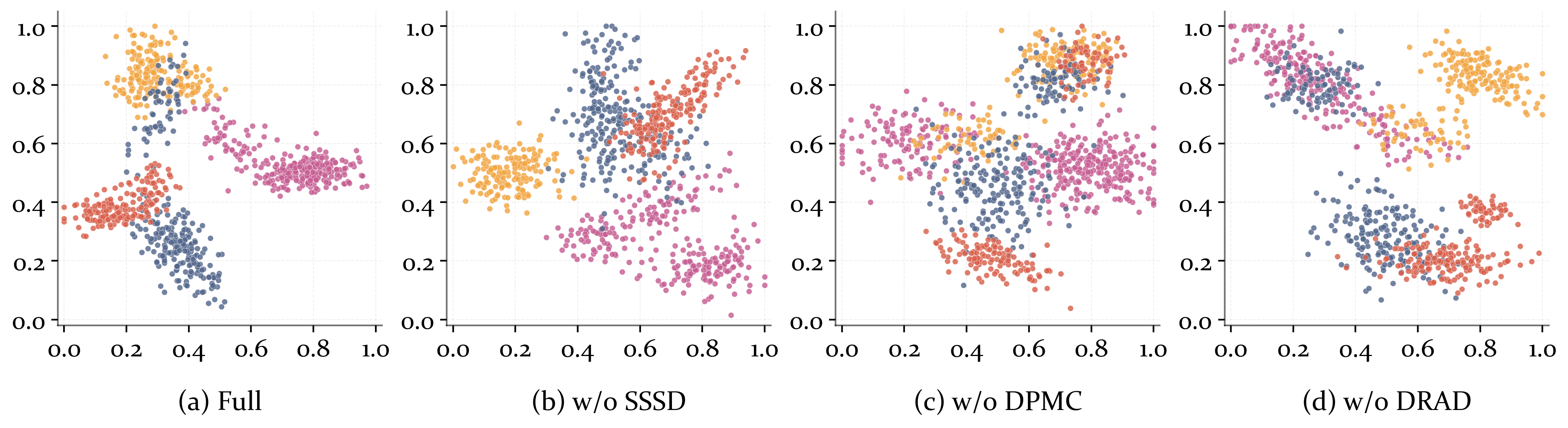} 
\caption{t-SNE visualization of fused representations on Four-Class IEMOCAP using only the text modality.}
\label{Figure9}
\end{figure*}

\begin{figure}[!t]
    \centering
    \includegraphics[width=\columnwidth]{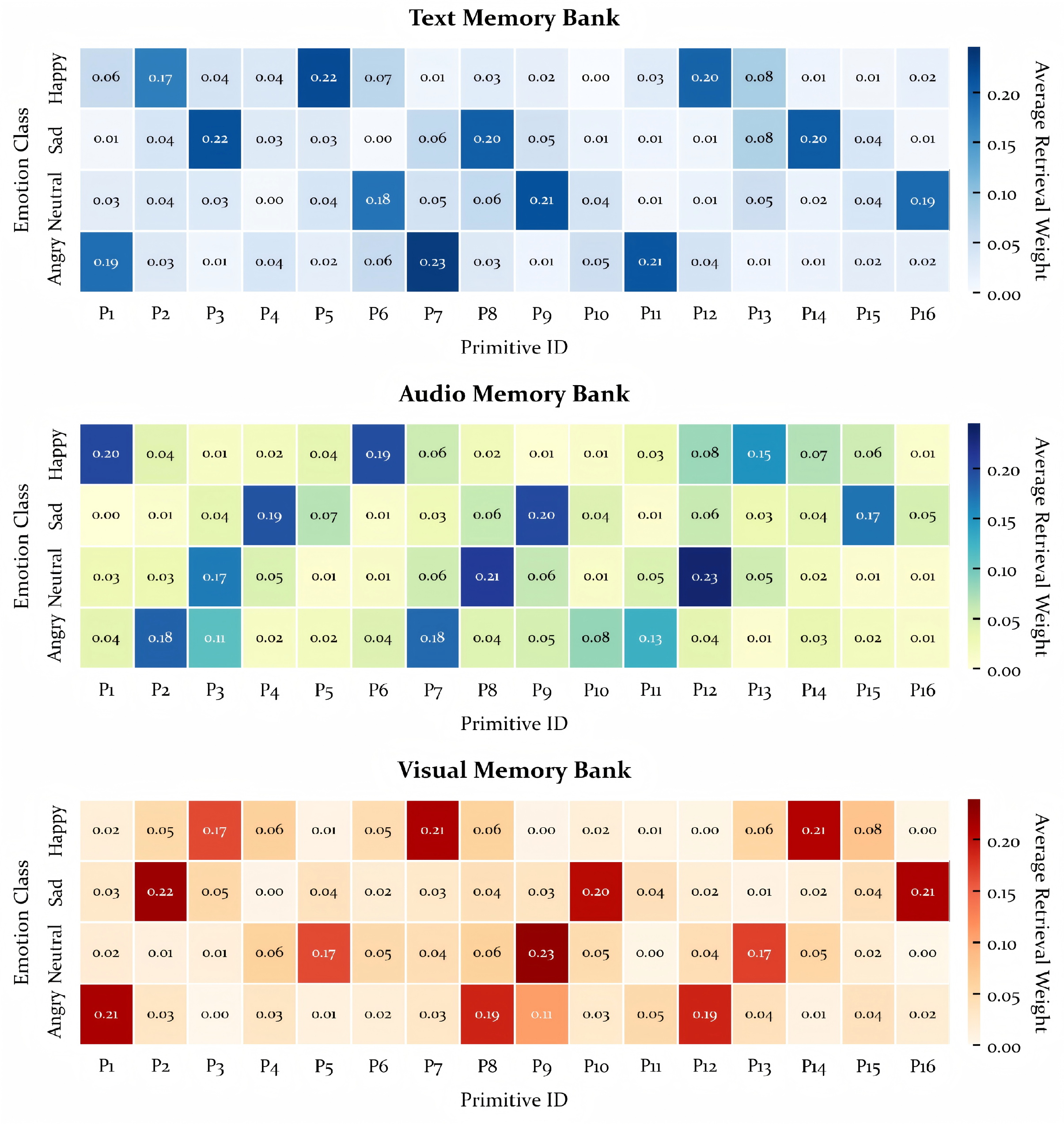}
    \caption{Visualization of class-level primitive retrieval patterns on audio, text, and visual memory banks.}
    \label{Figure8}
\end{figure}

\section{Visualization Analysis}

\label{app:visualization}

\subsection{Shared-Specific Relations}

To analyze the representation disentanglement effect of SSSD, we compute the average cosine similarity between \(S_A, S_T, S_V\) and \(P_A, P_T, P_V\). As shown in Figure~\ref{Figure7}, the three shared representations have high similarity, indicating that the shared semantics across different modalities are well aligned. In contrast, the similarity between shared and specific representations is clearly lower, suggesting that they are effectively separated in the representation space. Meanwhile, the specific representations of different modalities maintain moderate similarity, which indicates that they still preserve their modality-specific information and do not completely overlap. These results demonstrate that SSSD can achieve cross-modal alignment of shared semantics while retaining modality-specific information.

\subsection{Primitive Retrieval Patterns}

To analyze the retrieval behavior of the primitive memory, we visualize the average primitive retrieval weights of different 
emotion categories on each modality-specific memory bank in the test set. As shown in Figure~\ref{Figure8}, the horizontal axis denotes the primitive ID, and the vertical axis denotes the 
emotion category. The results show that different 
emotion categories tend to activate different subsets of primitives within the same memory bank. For example, Happy, Sad, Neutral, and Angry correspond to distinct high-weight regions. This indicates that the retrieval process does not randomly select primitives, but is correlated with the 
emotion 
category of the sample. Meanwhile, the activation patterns of the audio, text, and visual memory banks are not identical, suggesting that the memory banks of different modalities learn their own modality-specific primitive distributions. In addition, the retrieval weights are not concentrated on a few fixed primitives. Instead, they form category-related usage patterns across multiple primitives. This suggests that the primitive memory constructed by DPMC has certain diversity and can provide differentiated modality-specific information compensation for different emotion categories.

\subsection{Fused Representations}

To further analyze the effectiveness of different modules, we conduct t-SNE visualization on the IEMOCAP test set using only the text modality, comparing the classification representations of the full model and its ablated variants. As shown in Figure~\ref{Figure9}, we use pink, orange, gray, and yellow to denote Neutral, Happy, Sad, and Angry, respectively. The representations learned by the complete PriMD exhibit a clearer class structure, where samples from the same class are more compact and overlaps between different classes are reduced. In contrast, after removing SSSD, the sample distributions of different classes become more scattered, and clear mixing appears between some classes. After removing DPMC, the class boundaries become less distinct. After removing DRAD, the sample distribution becomes more entangled. These results indicate that the modules in PriMD jointly improve the class separability of emotion representations.

\end{document}